\documentclass[runningheads]{llncs}
\usepackage{subfigure}

\usepackage{amsthm}
 
\usepackage[final]{eccv}

\usepackage{eccvabbrv}

\usepackage{algorithm}
\usepackage{algorithmic}
\usepackage{times}
\usepackage{epsfig}
\usepackage{amssymb,enumitem}
\usepackage{graphicx}
\usepackage{amsmath}
\usepackage{booktabs}
\usepackage{array,multirow}
\usepackage{cite}
\usepackage{color}
\usepackage{xcolor}
\usepackage{makecell}
\usepackage{epsfig}
\usepackage{enumerate}
\usepackage{graphicx}
\usepackage{booktabs}

\usepackage[accsupp]{axessibility}  

\usepackage[pagebackref,breaklinks,colorlinks,citecolor=eccvblue]{hyperref}

\usepackage{orcidlink}

\begin{document}

\title{Rethinking Real-world Image Deraining via An Unpaired Degradation-Conditioned Diffusion Model} 

\titlerunning{Abbreviated paper title}

\author{Yiyang Shen\inst{1,3} \and
Mingqiang Wei\inst{1}\and
Yongzhen Wang\inst{1}\and
Xueyang Fu\inst{2}\and
Jing Qin\inst{3}}


\institute{Nanjing University of Aeronautics and Astronautics \and
University of Science and Technology of China \and
The Hong Kong Polytechnic University}

\maketitle

\begin{abstract}
Recent diffusion models have exhibited great potential in generative modeling tasks. Part of their success can be attributed to the ability of training stable on huge sets of paired synthetic data. However, adapting these models to real-world image deraining remains difficult for two aspects. First, collecting a large-scale paired real-world clean/rainy dataset is unavailable while regular conditional diffusion models heavily rely on paired data for training. Second, real-world rain usually reflects real-world scenarios with a variety of unknown rain degradation types, which poses a significant challenge for the generative modeling process. To meet these challenges, we propose \textbf{RainDiff}, the first real-world image deraining paradigm based on diffusion models, serving as a new standard bar for real-world image deraining. We address the first challenge by introducing a stable and non-adversarial unpaired cycle-consistent architecture that can be trained, end-to-end, with only unpaired data for supervision; and the second challenge by proposing a degradation-conditioned diffusion model that refines the desired output via a diffusive generative process conditioned by learned priors of multiple rain degradations. Extensive experiments confirm the superiority of our RainDiff over existing unpaired/semi-supervised methods and show its competitive advantages over several fully-supervised ones.
\end{abstract}

\begin{figure*}[htbp]
\centering
\subfigure[Real rainy image]{
\includegraphics[width=3.0cm]{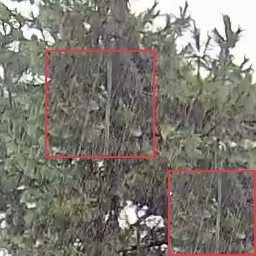}
}
\subfigure[CycleGAN]{
\includegraphics[width=3.0cm]{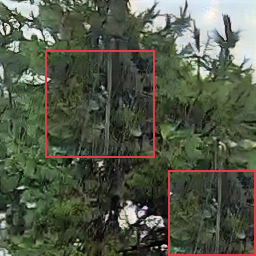}
}
\subfigure[DerainCycleGAN]{
\includegraphics[width=3.0cm]{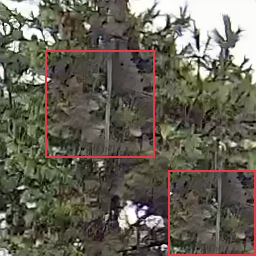}
}
\subfigure[DCD-GAN]{
\includegraphics[width=3.0cm]{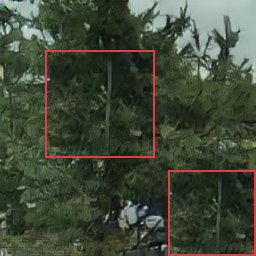}
}
\subfigure[NLCL]{
\includegraphics[width=3.0cm]{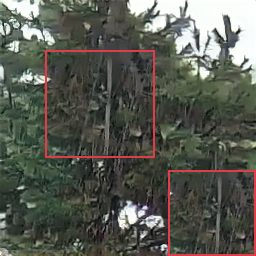}
}
\subfigure[Ours]{
\includegraphics[width=3.0cm]{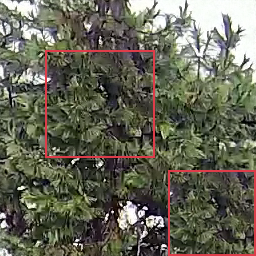}
}
\caption{Image deraining results on a real-world rainy image. From (a) to (f): (a) the real-world rainy image, the deraining results of (b) CycleGAN \cite{zhu2017unpaired}, (c) DerainCycleGAN \cite{wei2021deraincyclegan}, (d) DCD-GAN \cite{chen2022unpaired}, (e) NLCL \cite{ye2022unsupervised} and (f) our RainDiff. RainDiff generates both rain-free and perceptually more pleasing results.}
\label{ab_reals}
\end{figure*}

\section{Introduction}

Image deraining is an ill-posed problem. 
By learning from massive synthetic clean/rainy image pairs, the performance of learning-based techniques is substantially improved \cite{yang2017deep,fu2021rain}, compared to the traditional prior wisdom, such as sparse coding \cite{luo2015removing}, Gaussian Mixture Model \cite{li2016rain}, and low-rank representation \cite{chang2017transformed}. 
Despite their successes, these fully supervised methods achieve sub-optimal performance on real-world rainy images, because of i) the existence of the domain gap between synthetic and real-world rainy images, and ii) the difficulty to collect large-scale real-world clean/rainy image pairs.

To alleviate the aforementioned problems, semi-supervised deraining techniques, 
leverage paired synthetic data for good initialization and unpaired real-world data for generalization \cite{wei2019semi,ye2021closing,yasarla2020syn2real}. But the transferability is still limited since the rain patterns of synthetic images are fixed while the rain patterns of real-world images are dynamically changing.
%
Furthermore, the introduction of CycleGAN \cite{zhu2017unpaired} makes Generative Adversarial Networks (GANs) the preferred model family for real-world image deraining tasks, as they avoid the need for paired data \cite{wei2021semi,han2020decomposed,jin2019unsupervised,zhu2019singe,chen2022unpaired}.
%
However, these unpaired deraining methods are known as being difficult to train due to their complex adversarial objectives. As a result, they are susceptible to a series of problems, such as premature convergence, model collapse, and optimization instability. Moreover, these methods, limited only to single rain degradation cases, may not be optimal for multiple degradations in real-world rainy conditions, leading to image degradation including the loss of image details, remnant rain, halo artifacts, and/or color distortion. 

More recently, diffusion models \cite{sohl2015deep,ho2020denoising,song2019generative} have garnered significant attention for their effectiveness in a wide range of generative modeling tasks, such as image inpainting \cite{nichol2022glide}, image restoration \cite{ozdenizci2022restoring}, and image super-resolution \cite{saharia2022image}.
%
Compared to GANs, diffusion models offer a stable training process and exhibit greater efficacy in modeling the pixel distribution of images. However, no work to-date explores what will happen when unpaired learning meets diffusion models for real-world image deraining. We identify two major obstacles to their practical application in real-world image deraining. First, real-world rainy images lack corresponding clean images, which poses a challenge for existing diffusion models that typically prioritize synthetic degradation scenarios, where generating large-scale paired synthetic data is easier than for real-world examples. Models trained on paired synthetic data struggle to effectively handle unpaired real-world data. 
Second, real-world rain presents diverse degradation types, including but not limited to rain streaks, raindrops, rainy haze, and a combination of them (also called the mixture of rain), which usually changes over time, especially in heavy rain conditions. Thus, models primarily designed for single-degradation image processing may not be well generalized to
multiple real-world rain degradations.

We propose \textbf{RainDiff}, a new standard bar for real-world image deraining. RainDiff utilizes an effective unpaired cycle-consistent architecture with a degradation-conditioned diffusion model that achieves the desired deraining results under real-world rain scenarios with multiple rain degradations (see Fig. \ref{ab_reals}).

\textit{Why is the unpaired cycle-consistent architecture?} The unpaired cycle-consistent architecture fully takes advantage of the circulatory architecture to overcome the challenge of training without paired data. Instead of popular adversarial learning architectures for unpaired data, the proposed method offers a stable and non-adversarial training process that better facilitates real-world rain removal. 

\textit{Why is the degradation-conditioned diffusion model?} The degradation-conditioned diffusion model adds additional degradation-conditioned controls to the diffusion model, making it possible to handle diverse rain degradations. Such degradation-conditioned controls precisely express the space differences among various types of rain degradations, that enable finer diffusive generative processes in real-world rain conditions.

\textit{Why does RainDiff serve as a new standard bar for real-world image deraining?} RainDiff is an implementation of the idea of ``solving real-world image deraining in an unpaired learning manner '', such as DerainCycleGAN \cite{wei2021deraincyclegan} and DCD-GAN \cite{chen2022unpaired}. At the macro level, it is the first time to form an idea of applying the popular diffusion model to real-world image deraining; it solves several challenges encountered during the practical application process.
At the micro level, RainDiff introduces stable training of unpaired real-world data, rather than weakly adversarial training; It also learns priors of multiple rain degradations to enhance its performance in real-world deraining.

Extensive experiments show that RainDiff outperforms existing unpaired/semi-supervised methods and achieves comparable performance against fully-supervised ones. Overall, our contributions are as follows:
\begin{itemize}
\item We propose a novel unpaired learning paradigm via a degradation-conditioned diffusion model, called RainDiff, to generate quality real-world deraining results.
\item We propose an unpaired cycle-consistent architecture to provide a non-adversarial training process for unpaired data, where rain-related and clean-cue features can facilitate rain removal.
\item We propose a degradation-conditioned diffusion model to provide powerful diffusive progress for image deraining, where the learned priors of multiple rain degradations boost the generalization of deraining for diverse real-world rain scenarios.
\end{itemize}

\section{Related Work}

\textbf{Single Image Deraining.} 
Images captured under complicated rainy scenarios inevitably suffer from the noticeable degradation of visual quality. This degradation causes detrimental impacts on many vision tasks, including  segmentation \cite{liu2020deep}, object detection \cite{wojna2019devil}, and video surveillance \cite{sultani2018real}. Thus, it is indispensable to develop effective algorithms to recover quality rain-free images, which is referred to as \textit{image deraining}.
Early single image deraining methods employ hand-crafted priors, such as low-rank representation \cite{chen2013generalized,chang2017transformed}, sparse coding \cite{gu2017joint,wang2017hierarchical}, and Gaussian mixture model \cite{li2016rain}, to restore rainy images.
Recently, deep learning based methods have been substantiated to be effective in image deraining \cite{fu2017clearing}. The pioneering work \cite{fu2017clearing} introduces an end-to-end residual convolutional neural network (CNN) for simplifying the learning process.
Network modules, such as dense block \cite{li2018non,wang2019erl}, recursive block \cite{fan2018residual,li2018recurrent} and dilated convolution \cite{deng2020detail}, and structures, such as RNN \cite{li2018recurrent,ren2019progressive}, GAN \cite{zhang2019image,chen2022unpaired} and multi-stream networks \cite{yang2017deep,zhang2018density}, are validated to be effective in image deraining.
Despite the promising deraining results on synthetic datasets, these methods trained on such synthetic images generalize poorly to real-world images, typically because of the obvious domain gap between synthetic and real-world rainy images. To solve this issue, several semi-supervised frameworks have been proposed \cite{wei2019semi,ye2021closing,yasarla2020syn2real} to achieve improved generalization performance. However, these supervised/semi-supervised methods still require paired data, which is challenging or even impossible to obtain in real-world rainy scenes.
%
Motivated by the success of CycleGAN \cite{zhu2017unpaired}, a popular image-to-image translation architecture, recent works \cite{zhu2019singe,jin2019unsupervised,wei2021deraincyclegan,chen2022unpaired} attempt to exploit the improved CycleGAN architecture and constrained transfer learning to jointly learn the rainy and rain-free image domains.
However, these unpaired deraining methods heavily rely on complex adversarial objectives to develop their algorithms, which makes it difficult to achieve stable training. Additionally, they are limited in their ability to handle specific degradations.

\textbf{Denoising Diffusion Probabilistic Models.} Recently, denoising diffusion probabilistic models (DDPM) \cite{sohl2015deep,ho2020denoising,song2019generative} have exhibited their powerful ability in various vision tasks, such as image super-resolution \cite{saharia2022image}, text-to-image generation \cite{gu2022vector}, and image segmentation \cite{amit2021segdiff}. 
More recently, \"{O}zdenizci et al.~\cite{ozdenizci2022restoring} propose patch-based denoising diffusion models to demonstrate how diffusion models can be used for image restoration. However, these architectures have not been applied to real-world image deraining. They still require paired data for training and are designed for a specific degradation only. This observation serves as a motivation for us to propose the first diffusion-based model for real-world image deraining.


\section{RainDiff}\label{proposedmethod}
We start by discussing the necessary background and notation on diffusion models in Sec. \ref{bgdm}, and then we introduce our method in Sec. \ref{cyctitle} and Sec. \ref{dcdmtitle}.


\subsection{Denoising Diffusion Probabilistic Models}\label{bgdm}
Denoising diffusion probabilistic models (DDPM) slowly corrupt the training data with Gaussian noise and learn to reverse this corruption as a generative model \cite{sohl2015deep,ho2020denoising,song2019generative}. In the forward process, Gaussian noise is added sequentially onto an input image $x_{0} \sim q(x_{0})$ over $T$ time steps according to the Markovian process:
\begin{equation}
q(x_{1:T}|x_{0})= \prod_{t=1}^{T} q(x_{t}|x_{t-1})
\end{equation}
where $q(x_{t}|x_{t-1})= \mathcal{N}(x_{t};\sqrt{1-\beta_{t}}x_{t-1},\beta_{t}I)$, $\mathcal{N}(.)$ is  Gaussian distribution, $I$ is an identity covariance matrix with the same dimensions as the input image $x_{0}$. 
An important property of this forward process is its ability to directly sample any $x_{t}$ from $x_{0}$:
\begin{equation}
x_{t} = \sqrt{\overline{\alpha}_{t}}x_{0}+\epsilon_{t} \sqrt{1-\overline{\alpha}_{t}}
\end{equation}
where $\epsilon_{t} \sim \mathcal{N}(0,I)$, $\alpha_{t}=1-\beta_{t}$ and $\overline{\alpha}_{t}=\prod_{i=1}^{t}\alpha_{i}$. 
Similarly, reverse diffusion also adopts a Markov chain from $x_{T}$ onto $x_{0}$, albeit each step aims to gradually denoise the samples. Even though the reverse transition probability between $x_{t}$ and $x_{t-1}$ can be approximated as a Gaussian distribution under small $\beta_{t}$ and large $T$:
\begin{equation}
\label{reverse}
p_{\theta}(x_{t-1}|x_{t})= \mathcal{N}(x_{t-1};\mu_{\theta}(x_{t},t),\sigma_{\theta}(x_{t},t))
\end{equation}
\begin{equation}
\label{reverse_0:T}
p_{\theta}(x_{0:T})= p(x_{T}) \prod_{t=1}^{T} p_{\theta}(x_{t-1}|x_{t})
\end{equation}
where the reverse process is parameterized by a network to estimate $\mu_{\theta}(x_{t},t)$ and $\sigma_{\theta}(x_{t},t)$.  
Common parametrization focuses on $\mu_{\theta}(x_{t},t)$ while ignoring $\sigma_{\theta}(x_{t},t)$ \cite{ho2020denoising}:
\begin{equation}
\label{epsilon}
\mu_{\theta}(x_{t},t)= \frac{1}{\sqrt{\alpha_{t}}}(x_{t}-\frac{\beta_{t}}{\sqrt{1-\overline{\alpha}_{t}}}\epsilon_{\theta}(x_{t},t))
\end{equation}
In this case, the network is used to estimate the added noise $\epsilon_{t}$ by minimizing the loss:
\begin{equation}
\label{noise_loss}
L_{err}= \mathbb{E}_{x_{0},t,\epsilon_{t}\sim \mathcal{N}(0,I)}[||\epsilon_{t}- \epsilon_{\theta}(\sqrt{\overline{\alpha}_{t}}x_{0}+\epsilon_{t} \sqrt{1-\overline{\alpha}_{t}},t)||^{2}]
\end{equation}
During inference, reverse diffusion steps are performed starting from a random sample $x_{T} \sim \mathcal{N}(0,I)$. For each step $t \in \{T,...,1\}$, $\mu$ is derived by Eq. \ref{epsilon} based on the estimated $\epsilon_{\theta}$, and $x_{t-1}$ is sampled based on Eq. \ref{reverse} as:
\begin{equation}
\label{noise_loss2}
x_{t-1}=\frac{1}{\sqrt{\alpha_{t}}}(x_{t}-\frac{\beta_{t}}{\sqrt{1-\overline{\alpha}_{t}}}\epsilon_{\theta}(x_{t},t))+\sigma_{t}z
\end{equation}
where $z \sim \mathcal{N}(0,I)$) resembling one step of sampling via Langevin dynamics \cite{welling2011bayesian}.

To achieve high-quality image deraining, we need to learn a conditional reverse process $p_{\theta}(x_{0:T}|\Tilde{x})$ without modifying the diffusion process $q(x_{1:T}|x_{0})$ for $x$, where $x_{0}$ and $\Tilde{x}$ represent clean and rainy images, respectively. During the training phase, we sample $(x_{0},\Tilde{x}) \sim q(x_{0},\Tilde{x})$ from a paired data distribution and learn a conditional diffusion model. We input $\Tilde{x}$ to the reverse process as:
\begin{equation}
\label{reverse_rain}
p_{\theta}(x_{0:T}|\Tilde{x})= p(x_{T}) \prod_{t=1}^{T} p_{\theta}(x_{t-1}|x_{t},\Tilde{x})
\end{equation}
The noise estimators in Eqs. \ref{epsilon}-\ref{noise_loss2} are also replaced by $\epsilon_{\theta}(x_{t},\Tilde{x},t)$.

\begin{figure*}[!t] \centering
	\includegraphics[width=1\linewidth]{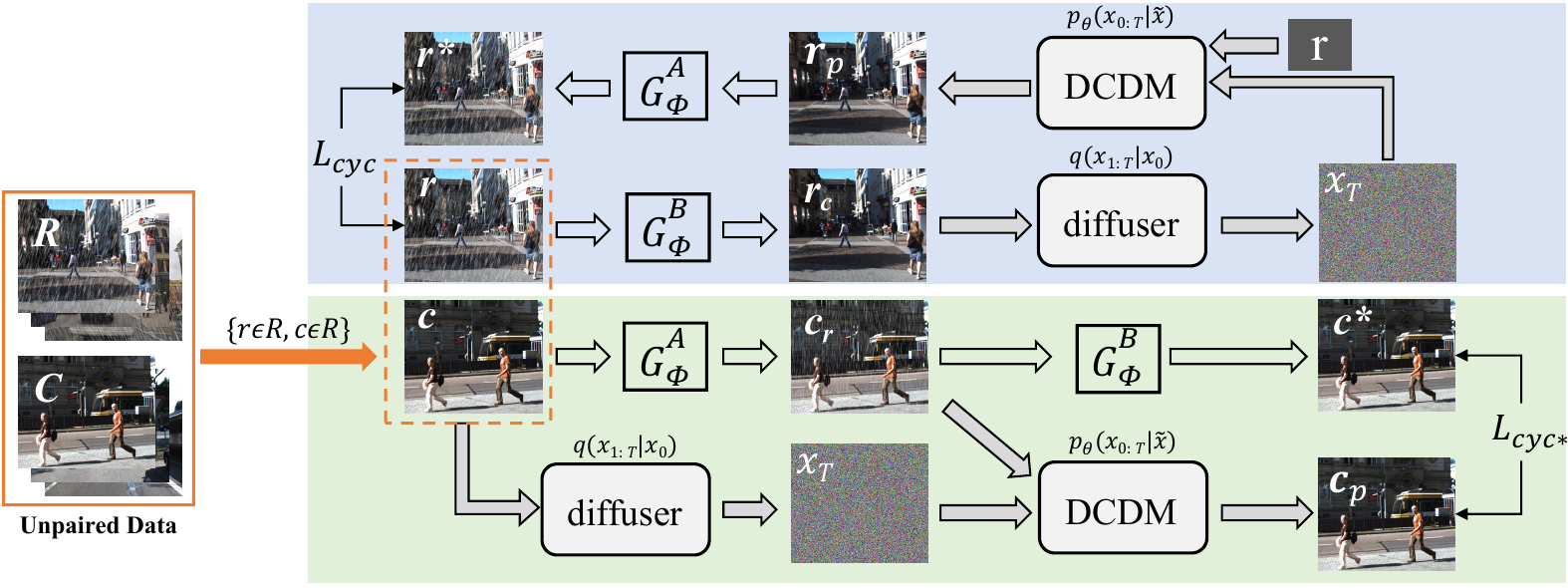}
	\caption{The pipeline of RainDiff. It takes unpaired clean/rainy data $\{C,R\}$ as input and trains an unpaired cycle-consistent architecture with a degradation-conditioned diffusion model (DCDM). Once trained, the model can produce high-quality real-world image deraining results, without access to paired clean images. Please refer to Sec. \ref{proposedmethod} for details.
    }
	\label{fig:network framework}
\end{figure*}
\subsection{Unpaired Cycle-consistent Architecture}\label{cyctitle}
Despite the impressive performance of diffusion models in image-conditional data synthesis and restoration \cite{ozdenizci2022restoring,saharia2022image}, their implementation still requires paired clean/rainy images.
We address this issue by designing a new unpaired cycle-consistent architecture without requiring adversarial training. It incorporates two cycle-consistent circuits with a degradation-conditioned diffusion model (DCDM) for unpaired training (see Fig. \ref{fig:network framework}).

Given an unpaired rainy image $\{r\in R\}$ and clean image $\{c \in C\}$, we first employ two non-diffusive generators with parameters $\phi^{A,B}$ to obtain the initial translation:
\begin{equation}\centering
\begin{split}
c_{r} = G_{\phi}^{A}(c), r_{c} = G_{\phi}^{B}(r)
\end{split}
\end{equation}
where $c_{r}$ and $r_{c}$ refer to the generated rainy image and rain-free image, respectively. Then, we use such clean/rainy image pairs to train DCDM for a conditional reverse process $p_{\theta}(x_{0:T}|\Tilde{x}), \Tilde{x}\in\{c_{r},r\}$ without modifying the diffusion process $q(x_{1:T}|x_{0}), x_{0}\in\{c,r_{c}\}$. Finally, we adopt two cycle-consistency loss functions to constrain the unpaired training procedure of the above cycle-consistent circuits:
\begin{equation}\label{cyc1}
L_{cyc} = E_{r\sim P_{data(r)}}[\left \| r^{*}-r \right \|_{1}
\end{equation}
\begin{equation}\label{cyc2}
L_{cyc^{*}} = E_{c_{p}\sim P_{data(c_{p})}}[\left \| c^{*}-c_{p} \right \|_{1}
\end{equation}
where $r^{*}=G_{\phi}^{A}(r_{p})$ and $c^{*}=G_{\phi}^{B}(c_{r})$, $c_{p}$ and $r_{p}$ refer to sampling results from the DCDM.
Especially, we adopt U-Net \cite{ronneberger2015u} as our non-diffusive generators. The training phase of RainDiff is outlined in Algorithm \ref{alg:training}.

\textbf{Difference between Existing Circulatory Structures and Ours.} In unpaired learning, the circulatory structures with cycle-consistency loss functions are commonly used for model training \cite{chen2022unpaired,wei2021deraincyclegan,zhu2017unpaired}. The differences between existing circulatory structures and ours lie in two aspects: 1) RainDiff requires no discriminators for adversarial training and provides a reliable non-adversarial training process. 2) It involves a degradation-conditioned diffusion model, which is primarily designed for image deraining. Unlike generators in existing circulatory structures, our $G_{\phi}^{A}$ and $G_{\phi}^{B}$ are both only used in the training phase and remain uninvolved in the testing phase for image deraining.

\begin{figure*}[!t]
	\centering
	\includegraphics[width=0.8\linewidth]{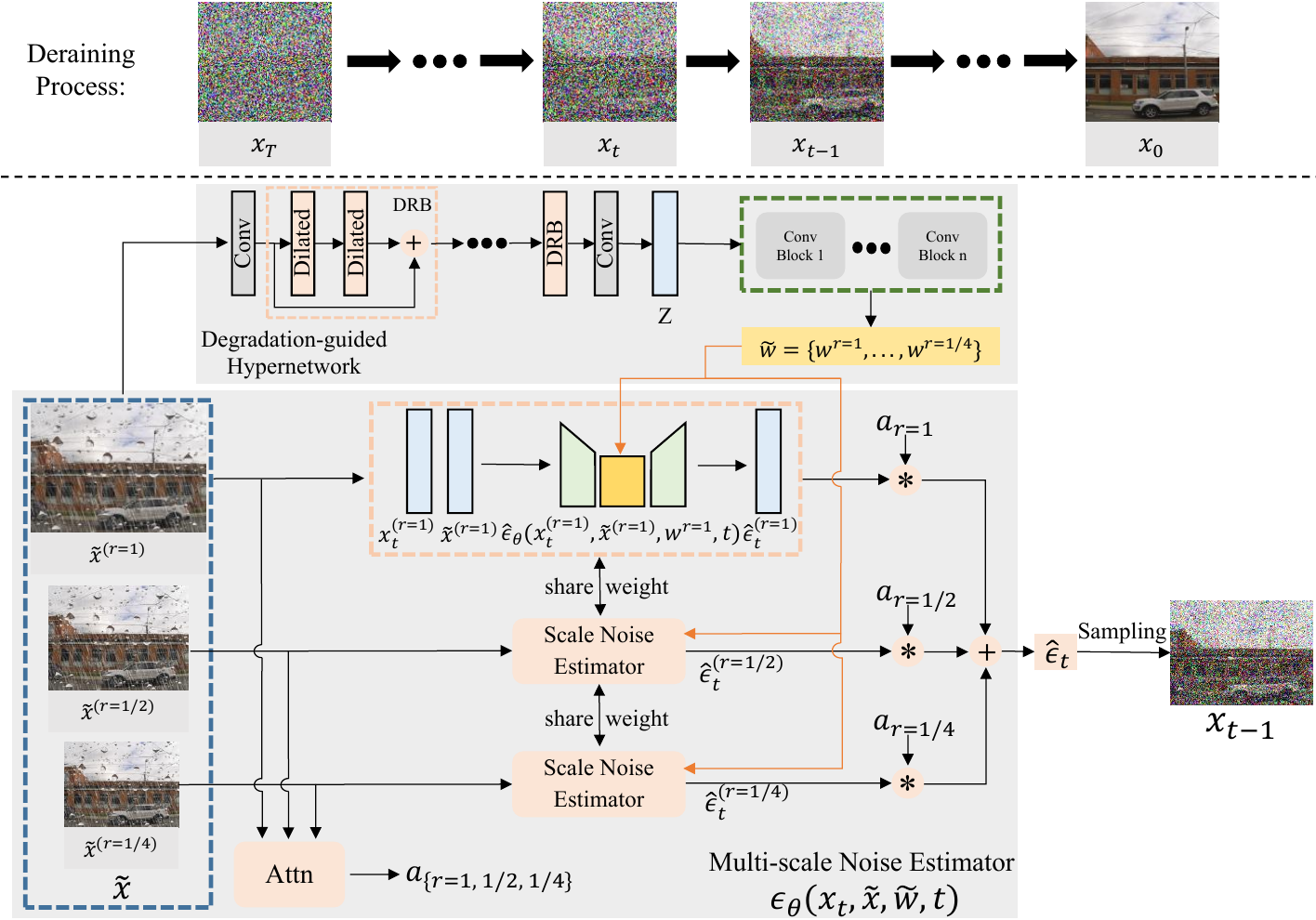}
    \caption{The degradation-conditioned diffusion model.}
	\label{fig_MDM}
\end{figure*}

\subsection{Degradation-Conditioned Diffusion Model}\label{dcdmtitle}

In the real world, rain exhibits multiple degradations, e.g., rain streaks, raindrops, rainy haze, and the mixture of rain, which may change over time. It presents a considerable challenge in directly implementing diffusion models to perform real-world image deraining. Beyond existing fixed-degradation diffusion models, we design a degradation-conditioned diffusion model (DCDM) to address this problem (see Fig. \ref{fig_MDM}). 

\textbf{Degradation-guided Hypernetwork.} As shown in Fig. \ref{fig_MDM}, our idea is to design a degradation-guided hypernetwork (DG-Hyper) that controls the reverse diffusion process using the learned degradation representation. Given a rainy image $\Tilde{x}$, we first feed it into $f_{D}$ to learn the latent degradation representation $Z = f_{D}(\Tilde{x})$, where $f_{D}$ represents two convolutional layers with five dilated residual blocks (DRBs) \cite{yu2017dilated} that helps enlarge the receptive field to capture more comprehensive characteristic of different rain degradations. We set the dilation rates of these five DRBs as $\{1,2,4,2,1\}$.

As a single type of conditional control for the diffusion model, $Z$ should possess adaptability to different rain degradations. It means that $Z$ will dynamically change in accordance with the variations in types of rain degradations, even with the same image contents. 
Inspired by contrastive learning \cite{wu2021contrastive,hadsell2006dimensionality,li2022all,he2020momentum}, we leverage a contrastive loss function to make $Z$ pull the representation with the same degradation $Z^{+}$ (called positive samples), and push apart the representation between negative samples $Z^{-}$ with different degradations. The proposed contrastive loss $L_{cl}$ is reformulated as:
\begin{equation}
L_{cl} = \sum_{i=1}^{n}\omega_{i} \bullet \frac{\mathop{ {\parallel} } \varphi_{i}(Z^{+})- \varphi_{i}(Z) \mathop{ {\parallel} }^{2}}{\mathop{ {\parallel} } \varphi_{i}(Z^{-})- \varphi_{i}(Z) \mathop{ {\parallel} }^{2}},
\end{equation}
where $\varphi_{i}(\cdot)$, $i = 1,2,...,n$, refers to extracting the $i$-th hidden features from the pre-trained VGG-16 network. We choose the 2-nd, 3-rd, and 5-th max-pooling layers. Similar to \cite{wu2021contrastive}, $\omega_{i}$ are weight coefficients with $\omega_{1}=0.2$, $\omega_{2}=0.5$, and $\omega_{3}=1$. Note that $Z^{+}$ and $Z^{-}$ are generated by feeding patches from $\Tilde{x}$ and other images, respectively.

Based on the learned latent degradation representation $Z$, we aim to enhance the adaptability of diffusion models to multiple rain degradations. To achieve it, we use a meta-learning technique, often referred to as hypernetwork \cite{45803}, to control the diffusion model using $Z$. Especially, we input $Z$ into a series of convolution blocks, consisting of a 1×1 convolutional layer with channel
groups, to generate the part of kernels' parameters (weights) $\Tilde{w}=\{w^{r=1},...,w^{r=1/4}\}$ of the scale noise estimators in the multi-scale noise estimator (primary network) $\hat{\epsilon}_{\theta} (x_{t}^{r},\Tilde{x}^{r},w^{r},t)$. Notably, if U-Net \cite{ronneberger2015u} is employed as the scale noise estimator, only two up-convolution and down-convolution layers at the highest resolution are optimized using these kernel weights. In summary, the introduction of DG-Hyper that learns to modulate the weights of the noise estimator in order to accurately represent latent degradation enables enhanced generalization of the diffusion model across various rain degradations.

\textbf{Multi-scale Noise Estimator.} We observe that rain degradations, particularly rain streaks, exhibit a wide range of characteristics, including but not limited to varying directions, densities, and sizes. These diverse rain patterns demonstrate minimal variation across different scales. Consequently, we incorporate this correlation of rain across multiple scales into the diffusion generative process to develop a multi-scale noise estimator for image deraining (see Fig. \ref{fig_MDM}). 

At each time step $t$, when provided with an intermediate sample $x_{t}$ and a rainy image $\Tilde{x}$, our initial step involves downsampling the original images into various scales, such as 1/2 and 1/4, as represented by:
\begin{equation}\label{eq:down}
\begin{aligned}
\{x_{t}^{r},\Tilde{x}^{r}\} = down(x_{t},\Tilde{x}),r\in\{1,1/2,1/4\}
\end{aligned}
\end{equation}
where $r$ represents the scale of the image. To make full use of multi-scale information for deraining, we then use attention heads \cite{chen2016attention} to learn all attention masks $\alpha_{r=\{1,1/2,1/4\}}$ for each of a fixed set of scales, which can be used to weight the multi-scale features at each pixel location. 
For each scale branch, the scale noise estimator $\hat{\epsilon}_{\theta}$ receives the intermediate variable $x_{t}^{r}$ and rainy image $\Tilde{x}^{r}$ from a single (lower) scale $r$ along with the time step $t$ as input to predict the noise map $\hat{\epsilon}^{r}_{t}$, and its part of kernels' weights are optimized with $w^{r}$ from DG-Hyper:
\begin{equation}\label{eq:noise_mul}
\begin{aligned}
\hat{\epsilon}^{r}_{t}=\hat{\epsilon}_{\theta}(x_{t}^{r},\Tilde{x}^{r},w^{r},t)
\end{aligned}
\end{equation}
We combine the noise maps from multiple scales, i.e., $\hat{\epsilon}^{r\in\{1,1/2,1/4\}}_{t}$, by multiplying the attention masks $\alpha_{r=\{1,1/2,1/4\}}$ with the maps in a pixel-wise manner, and then summing the results across different scales to obtain the final noise map $\hat{\epsilon}_{t}$. The whole multi-scale noise estimator $\epsilon_{\theta}(x_{t},\Tilde{x},\Tilde{w},t)$ is expressed as:
\begin{equation}\label{eq:noise_finals}
\begin{aligned}
\epsilon_{\theta}(x_{t},\Tilde{x},\Tilde{w},t)=\hat{\epsilon}_{t}=\sum^{N}_{i=1} a_{r} \ast \hat{\epsilon}^{r}_{t}
\end{aligned}
\end{equation}
where $N$ denotes the total number of different scales $r$. Especially, our multi-scale noise estimator incorporates multi-scale rain information into the diffusion generative process and introduces an attention mechanism that softly weights the multi-scale features at each pixel location. 

\begin{algorithm}[!t]
    \caption{The training of RainDiff}
    \label{alg:training}
    \begin{algorithmic}[1]
    \REQUIRE 
    Unpaired clean image $x$, rainy image $y$.
        \REPEAT
        \STATE $c_{r} = G_{\phi}^{A}(c), r_{c} = G_{\phi}^{B}(r)$\\
        \STATE $t \sim$ Uniform $(\{ 1,...,T \})$\\
        \STATE $\epsilon \sim \mathcal{N}(0,I)$\\
        \STATE $x_{T} = q(x_{1:T}|x_{0}), x_{0}\in\{c,r_{c}\}$\\
        \STATE $\{c_{p},r_{p}\}=p_{\theta}(x_{0:T}|\Tilde{x}), \Tilde{x} \in\{c_{r},r\}$ \\
        \STATE $c^{*} = G_{\phi}^{B}(c_{r}), r^{*} = G_{\phi}^{A}(r_{p})$\\
        \STATE Take gradient descent step on\\
        \STATE $\nabla_{\theta,\phi}[||\epsilon- \epsilon_{\theta}(\sqrt{\overline{\alpha}_{t}}c+\epsilon_{t} \sqrt{1-{\overline{\alpha}_{t}}},c_{r},\Tilde{w}_{t},t)||^{2}$\\
        \STATE $+\nabla_{\theta,\phi}[||\epsilon- \epsilon_{\theta}(\sqrt{\overline{\alpha}_{t}}r_{c}+\epsilon_{t} \sqrt{1-{\overline{\alpha}_{t}}},r,\Tilde{w}_{t},t)||^{2}$\\
        \STATE $+\lambda_{cyc}L_{cyc}+\lambda_{cyc^{*}}L_{cyc^{*}}+\lambda_{cl}L_{cl}]$
        \UNTIL{converged}\\
    \STATE \textbf{return} $\theta$
    \end{algorithmic}
\end{algorithm}

\begin{algorithm}[!t]
    \caption{The testing of RainDiff}
    \label{alg:test}
    \begin{algorithmic}[1]
    \REQUIRE 
    Rainy image $\Tilde{x}$, multi-scale noise estimator $\epsilon_{\theta}(x_{t},\Tilde{x},\Tilde{w},t)$, number of implicit sampling iterations $T$, and DG-Hyper $f(.)$.
    \STATE $x_{t} \sim \mathcal{N}(0,I)$\\
    \STATE $\Tilde{w}= f(\Tilde{x})$\\
    \FOR{each $i= S,...,1$}
        \STATE $t = (i-1)\cdot T/S+1$
        \STATE $t_{next} = (i-2)\cdot T/S+1$ \textbf{if} $i>1$ \textbf{else} 0
        \STATE $\hat{\epsilon}_{t}=\epsilon_{\theta}(x_{t},\Tilde{x},\Tilde{w},t)$
        \STATE $x_{t-1} = \sqrt{\overline{\alpha}_{t_{next}}}(\frac{x_{t}-\sqrt{1-\overline{\alpha}_{t}}\cdot\hat{\epsilon}_{t}}{\sqrt{\overline{\alpha}_{t}}})+\sqrt{1-\overline{\alpha}_{t_{next}}}\cdot \hat{\epsilon}_{t}$
    \ENDFOR
    \STATE\textbf{return} $x_{0}$
    \end{algorithmic}
\end{algorithm}

\begin{table*}[!t]
\centering
\caption{Descriptions of the established mixture dataset.}
\footnotesize
\label{datasets}
\begin{tabular}{c|c|c|c|c|c|c}
\toprule
\multirow{2}{*}{Type} & \multicolumn{6}{c}{Synthetic}\\                         
\cline{2-7} & RS     & RH     & RD     & RDS    & RHS            & Total\\ \midrule
Train-Set             & RainDS & OTS & RainDS & RainDS & RainCityscapes & -\\
Num & 1000& 1000& 1000& 1000& 1000& 5000\\ 
Test-Set              & RainDS & OTS & RainDS & RainDS & RainCityscapes & -\\
Num & 200& 200& 200& 200& 200& 1000\\\bottomrule
\multirow{2}{*}{Type} & \multicolumn{6}{c}{Real-world}\\
\cline{2-7} & RS     & RH     & RD     & RDS    & RHS            & Total\\ \midrule
Train-Set             &RainDS & RTTS & RainDS & RainDS & GT-Rain & - \\
Num&150&150&150&150&150&750\\ 
Test-Set              &RainDS & RTTS & RainDS & RainDS & GT-Rain & - \\ 
Num &98&98&98&98&98&490\\
\bottomrule
\end{tabular}
\end{table*}

The testing phase of RainDiff is outlined in Algorithm \ref{alg:test}. A large $T$ inevitably leads to costly sampling, e.g., when $T = 1000$. To address this problem, we use an implicit sampling strategy \cite{song2020denoising} to accelerate our sampling process (lines 4-5 in Alg. \ref{alg:test}). Implicit sampling with a multi-scale noise estimator $\epsilon_{\theta}(x_{t},\Tilde{x},\Tilde{w},t)$ can be performed by:
\begin{equation}
\label{implicit}
\begin{split}
x_{t-1}=&\sqrt{\overline{\alpha}_{t-1}}(\frac{x_{t}-\sqrt{1-\overline{\alpha}_{t}}\cdot \epsilon_{\theta}(x_{t},\Tilde{x},\Tilde{w},t)}{\sqrt{\overline{\alpha}_{t}}})\\
&+\sqrt{1-\overline{\alpha}_{t-1}}\cdot \epsilon_{\theta}(x_{t},\Tilde{x},\Tilde{w},t)
\end{split}
\end{equation}
during accelerated sampling we only needs a subsequence $\tau_{1},...,\tau_{S}$ of the complete ${1,...,T}$ timestep indices, which can be performed by:
\begin{equation}
\label{implicit2}
\begin{split}
\tau_{i}=(i-1) \cdot T / S+1
\end{split}
\end{equation}
At any denoising time step $t$, according to Eq. \ref{eq:noise_finals}, we utilize the multi-scale noise estimator $\epsilon_{\theta}(x_{t},\Tilde{x},\Tilde{w},t)$ from lines 9-10 in Alg. \ref{alg:training} to estimate the noise map $\hat{\epsilon}_{t}$ (line 6 in Alg. \ref{alg:test}). Subsequently, we perform an implicit sampling update utilizing the noise map $\hat{\epsilon}_{t}$ (line 7 in Alg. \ref{alg:test}).

\section{Experiment}
\subsection{Experimental Settings}
\textbf{Implementation Details.} We implement RainDiff using Pytorch 1.6 on an Nvidia GeForce RTX 3090 GPU. For optimizing RainDiff, we use the Adam optimizer with a min-batch size of 4 to train the paradigm, where the momentum parameters $ \beta_{1}$ and $\beta_{2}$ take the values of 0.5 and 0.999, respectively. 
The initial learning rate is set to $1e^{-4}$. For training, a 128 $\times$128 patch is randomly cropped from the original image (or its horizontal flipped version). The balance weights $\lambda_{cyc}$, $\lambda_{cyc^{*}}$ and $\lambda_{cl}$ are both set to 1.

\textbf{Datasets.} 
We adopt four challenging benchmark datasets to create a mixture dataset with five different rain degradations for evaluation (see Table \ref{datasets}), i.e., RainDS \cite{quan2021removing} includes synthetic and real-world images of rain streaks, raindrops, and the combination of them as well as their corresponding clean images, 
RESIDE-$\beta$ \cite{li2018benchmarking} includes synthetic and real-world hazy images (OTS set and RTTS set), 
RainCityscapes \cite{hu2019depth} and GT-Rain \cite{ba2022not} include synthetic and real-world images of the combination of rain streaks and rainy haze, respectively. 
Take a synthetic train set as an example, each train set of different rain degradations contains 1000 images to build a whole mixture train set (including 5000 images). Similarly, the total number of images of synthetic test sets and real-world train and test sets are 1000, 750, and 490, respectively.
To ensure a balanced representation of diverse rain degradations, we carefully select a specific number of images from RESIDE-$\beta$ \cite{li2018benchmarking}, RainCityscapes \cite{hu2019depth}, and GT-Rain \cite{ba2022not} based on the aforementioned criteria to construct the mixture dataset.
Notably, we also collect some real-world rainy images without ground truth from the Internet for testing.

\begin{table*}[!t]
\centering
\scriptsize
\caption{Comparisons of different deraining methods on the synthetic datasets. \textbf{Bold} and \underline{underline} indicate the best and second-best results. }
\label{synthetic_quantity}
\begin{tabular}{c|c|c|c|c|c|c}
\toprule
\multirow{3}{*}{Method} & \multirow{3}{*}{Type} & \multicolumn{5}{c}{Synthetic} \\
\cline{3-7} 
                        &                       & RS  & RH  & RD  & RDS  & RHS  \\ \cline{3-7} 
 &                       & PSNR/SSIM  & PSNR/SSIM  & PSNR/SSIM  & PSNR/SSIM  & PSNR/SSIM  \\                                \midrule
DSC \cite{luo2015removing}&Prior-based                 & 13.18/0.456    &17.46/0.727     &14.54/0.607     &13.49/0.482      &15.33/0.629      \\
GMM \cite{li2016rain}&Prior-based                   &14.64/0.485     &   18.21/0.718  &16.35/0.659      &15.92/0.585      &16.75/0.683      \\ \midrule
DDN \cite{detail_layer}&Supervised                  &23.79/0.692     &   22.42/0.882  &22.92/0.836      & 19.74/0.644     &18.66/0.790      \\
DID-MDN \cite{zhang2018density}&Supervised                 &25.26/0.758     &22.95/0.887     & 24.86/0.890    &21.56/0.718      &20.15/0.803      \\
SPA-Net \cite{wang2019spatial}&Supervised                 &31.09/0.906     &25.38/0.915     &27.98/\textbf{0.920}     &25.82/0.847      & 23.38/0.898  \\
DRD-Net \cite{deng2020detail} &Supervised                 & 29.97/0.893    &25.22/0.904     &28.15/0.907    &22.43/0.696      &21.49/0.864  \\
WeatherDiffusion \cite{ozdenizci2022restoring} &Supervised                 &\textbf{32.15}/\textbf{0.924}     &\underline{26.94}/\underline{0.922}     &\underline{30.22}/0.913     & \underline{28.09}/\textbf{0.882}     & \underline{23.84}/0.892  \\  \midrule                      
SIRR \cite{wei2019semi} &Semi-super                  &29.45/0.878     &  23.93/0.874   &25.93/0.897      &24.39/0.833      & 20.59/0.873     \\
Syn2Real \cite{yasarla2020syn2real}&Semi-super              &30.33/0.908    &24.19/0.918     &26.44/0.911     &26.97/0.858      &21.31/\underline{0.901}    \\
JRGR \cite{ye2021closing}&Semi-super                 &30.92/0.909     &  25.39/0.916   & 25.58/0.902    &26.76/0.864      &22.56/0.896      \\ \midrule
CycleGAN \cite{zhu2017unpaired}&Unpaired               &25.08/0.764     &23.22/0.911     &21.19/0.784    &20.58/0.663      &22.60/0.888      \\
DerainCycleGAN \cite{wei2021deraincyclegan}&Unpaired      &25.40/0.770     &20.14/0.812     &20.69/0.799      &20.06/0.660      &19.06/0.865         \\
DCD-GAN \cite{chen2022unpaired}&Unpaired                &22.80/0.737     &21.46/0.789     &21.43/0.689     &21.07/0.651      &22.90/0.829      \\
NLCL \cite{ye2022unsupervised} &Unpaired                 &24.12/0.808     &21.17/0.846     &22.37/0.832     &21.14/0.719      &22.45/0.891    \\
Ours &Unpaired                 &\underline{31.30}/\underline{0.911}     &\textbf{27.32}/\textbf{0.928}     & \textbf{30.54}/\underline{0.915}    &     \textbf{28.41}/\underline{0.878} &\textbf{24.28}/\textbf{0.906}    \\ \bottomrule
\end{tabular}

\end{table*}

\begin{table*}[!t]\footnotesize
\centering
\footnotesize
\caption{Comparisons of different deraining methods on the real-world datasets. \textbf{Bold} and \underline{underline} indicate the best and second-best results. }
\label{real-world_quantity}
\setlength{\tabcolsep}{0.25mm}{
\begin{tabular}{c|c|c|c|c|c}
\toprule
\multirow{3}{*}{Method} & \multirow{3}{*}{Type} &  \multicolumn{4}{c}{Real-world} \\
\cline{3-6}&                       &  RS     & RD  & RDS  & RHS  \\ \cline{3-6} 
 &                       & PSNR/SSIM  & PSNR/SSIM  & PSNR/SSIM  & PSNR/SSIM  \\                                \midrule
DSC \cite{luo2015removing}&Prior-based                 & 15.14/0.496     & 14.35/0.469    &13.51/0.443     & 14.98/0.517       \\
GMM \cite{li2016rain}&Prior-based                   & 17.48/0.527    &14.77/0.481     &14.24/0.475      &15.18/0.532      \\ \midrule
DDN \cite{detail_layer}&Supervised                  & 18.39/0.608    &17.23/0.525     & 15.67/0.522     & 17.41/0.590     \\
DID-MDN \cite{zhang2018density}&Supervised                 &19.74/0.645      &17.82/0.554     &17.36/0.563     & 18.98/0.617        \\
SPA-Net \cite{wang2019spatial}&Supervised                 &21.54/0.672      &18.95/0.563     & 19.12/0.608    &21.28/\underline{0.656}      \\
DRD-Net \cite{deng2020detail} &Supervised                 & 22.49/0.716      &19.34/0.636     &18.96/0.590     & 20.82/0.629        \\
WeatherDiffusion \cite{ozdenizci2022restoring} &Supervised                 &\underline{23.78}/\underline{0.739}     &20.57/0.648       &19.90/0.622      & 21.08/0.618     \\  \midrule                      
SIRR \cite{wei2019semi} &Semi-super                  &    23.42/0.718 &   20.66/0.610  & 17.98/0.576    & 21.04/0.630     \\
Syn2Real \cite{yasarla2020syn2real}&Semi-super              &23.17/0.691    &19.27/0.621        &18.54/0.583     &\underline{21.65}/0.642      \\
JRGR \cite{ye2021closing}&Semi-super                 &     23.49/0.712&  \underline{20.74}/\underline{0.652}   &18.13/0.560     &  20.39/0.618      \\ \midrule
CycleGAN \cite{zhu2017unpaired}&Unpaired               &21.07/0.612      &18.24/0.568     &17.55/0.515      &18.71/0.572      \\
DerainCycleGAN \cite{wei2021deraincyclegan}&Unpaired      &21.08/0.615       &18.43/0.565    &17.99/0.523      &  21.06/0.648    \\
DCD-GAN \cite{chen2022unpaired}&Unpaired                &18.81/0.692      &17.81/0.553     &\underline{20.74}/\underline{0.654}      &17.84/0.537      \\
NLCL \cite{ye2022unsupervised} &Unpaired                 &21.71/0.699      &19.24/0.606     &19.88/0.580      &20.18/0.603      \\
Ours &Unpaired                 &\textbf{24.73}/\textbf{0.770}   & \textbf{21.89}/\textbf{0.685}    &    \textbf{22.18}/\textbf{0.681}  &  \textbf{22.66}/\textbf{0.694}   \\ \bottomrule
\end{tabular}
}
\end{table*}
\textbf{Comparison Methods.} We qualitatively and quantitatively compare our method with two prior-based algorithms (i.e., DSC \cite{luo2015removing}, GMM \cite{li2016rain}), five paired supervised methods (i.e., DDN \cite{detail_layer}, DID-MDN \cite{zhang2018density}, SPA-Net \cite{wang2019spatial}, DRD-Net \cite{deng2020detail}, and WeatherDiffusion \cite{ozdenizci2022restoring}), three semi-supervised methods (i.e., SIRR \cite{wei2019semi}, Syn2Real \cite{yasarla2020syn2real}, and JRGR \cite{ye2021closing}), as well as four unpaired methods (i.e., CycleGAN \cite{zhu2017unpaired}, DerainCycleGAN \cite{wei2021deraincyclegan}, DCD-GAN \cite{chen2022unpaired} and NLCL \cite{ye2022unsupervised}).
Two popular metrics are used for quantitative comparisons, i.e., Peak Signal-to-Noise Ratio (PSNR) \cite{huynh2008scope} and Structure Similarity (SSIM) \cite{wang2004image}. Higher value of these metrics indicates better performance of the methods.
For fair comparisons, we re-train these methods on the mixture dataset that consists of the above five distinct rain degradations.

\begin{figure*}[!t]
	\centering
		\includegraphics[width=1\linewidth]{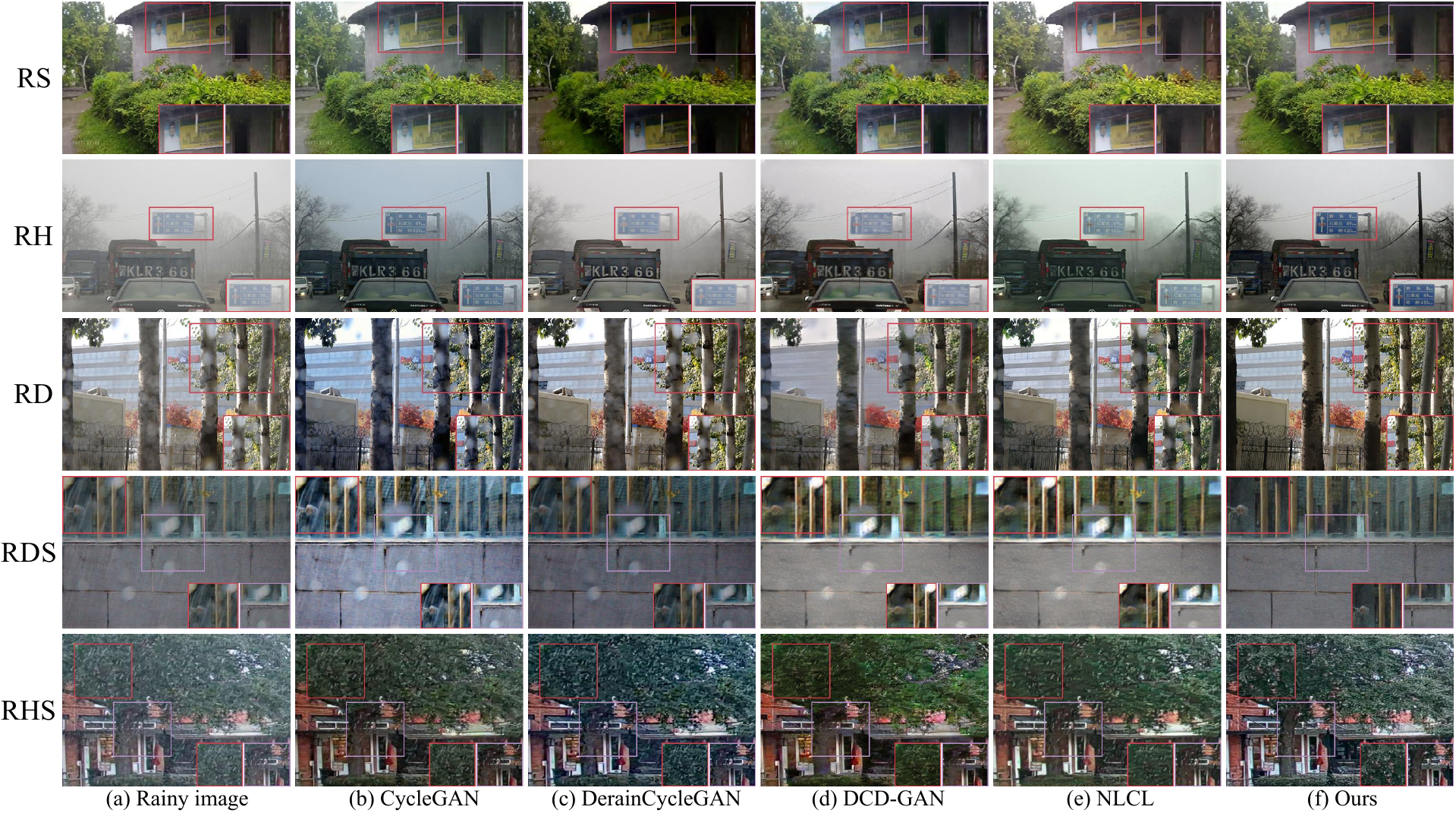}
    \caption{Comparison of deraining performance on real-world rainy images. Our method is more successful to remove different rain degradations and obtains the cleanest result with clear details.}
	\label{morecompare}
\end{figure*}

\subsection{Comparison with State-of-the-arts}
\textbf{Comparison on Synthetic Datasets.} 
Table \ref{synthetic_quantity} presents the quantitative results of different methods on five synthetic test sets. We make the following observations: 1) Compared with 
unpaired deraining methods, our method obtains higher values of PSNR and SSIM, which verifies the excellent performance of RainDiff.
2) There is an obvious performance gap between semi-supervised and supervised methods, which can be even more significant than the gap between RainDiff and supervised ones.
3) Our method is capable of achieving competitive results to existing supervised ones even without the paired data for supervision.

\begin{table*}[!t]
\centering
\scriptsize
\caption{Ablation study for different models on the real-world RDS test set. $L_{err}(c,c_{r})$ indicates line 9 in Alg. \ref{alg:training}.}
\label{loss_ab}
\begin{tabular}{cccccccccc}
\toprule
Model       & w/o $L_{err}(c,c_{r})$& w/o $L_{err}(r_{c},r)$& w/o $L_{cyc}$& w/o $L_{cyc}^{*}$&w/o Up&w/o Low&w/o DG-Hyper&Ours\\ \midrule
PSNR &20.66     &  20.98&19.57 &19.85&17.48&17.63&18.98&\textbf{22.18} \\
SSIM & 0.671    & 0.675&0.649 &0.654&0.534&0.547&0.634&\textbf{0.681} \\
\bottomrule
\end{tabular}
\end{table*}

\begin{table}[!t]
\footnotesize
\centering
\caption{Ablation analysis for different scales on the real-world RDS test set.}
\begin{tabular}{ccccc}
\toprule
Settings & V1                        & V2                        & V3                        & V4                        \\
$r=1$        & \checkmark & \checkmark & \checkmark & \checkmark \\
$r=1/2$      & w/o                       & \checkmark & w/o                       & \checkmark \\
$r=1/4$      & w/o                       & w/o                       & \checkmark & \checkmark \\ \midrule
PSNR     & 20.68 & 21.75                    & 21.46                                          & \textbf{22.18}                     \\
SSIM     & 0.658  & 0.679                   & 0.670                                          & \textbf{0.681} 
\\ \bottomrule
\end{tabular}
\label{scale}
\end{table}

\textbf{Comparison on Real-world Datasets.} 
For further general verification in practical use, we compare RainDiff with different methods on five real-world test sets. Notably, these test sets, except for the hazy set, both contain real-world rainy images along with their corresponding ground truth for evaluation using the numerical metrics. And we visualize the real-world haze removal results in Fig. \ref{morecompare}. Table~\ref{real-world_quantity} demonstrates that our method achieves superior performance gains to all compared methods. The incorporation of additional constraints provided by DG-Hyper enables RainDiff to effectively handle diverse rain degradations in real-world rainy images.

In Fig. \ref{morecompare}, we visualize the rain removal results on the collected real-world rainy images. RainDiff successfully handles diverse rain degradations and achieves superior visual results than all compared methods. Notably, the deraining results of unpaired methods may induce color and structure distortion, whereas our method can better preserve the color and structure of the image. More qualitative results are provided in the supplementary material.


\subsection{Ablation Study}
\label{Ablation_Study}

\textbf{Effect of Loss Function.} We evaluate the effectiveness of our hybrid loss function on the real-world RDS test set. 
Especially, we remove one component to each configuration at one time. For fair comparison, the same training settings are kept for all models testing.
As depicted in Table \ref{loss_ab}, the full structure of RainDiff exhibits the highest performance in both PSNR and SSIM metrics, suggesting that all the components of RainDiff are advantageous for proficient rain removal. 
%

%
\textbf{Effect of DG-Hyper.} To show the effectiveness of our DG-Hyper, we conduct an ablation study on the real-world RDS test set by removing DG-Hyper with the corresponding loss function $L_{cl}$. From Table \ref{loss_ab}, we can see that DG-Hyper can add an additional constraint for multiple rain degradations, which further improves the real-world deraining performance of the proposed method. Furthermore, we train it with different combinations of multiple degradations to analyze how the performance is influenced by different degradations. The results are provided in the supplementary material.

\textbf{Effect of Cycle-consistent Circuits.} 
As shown in Fig. \ref{proposedmethod}, there are two cycle-consistent circuits in the unpaired cycle-consistent architecture. To further validate their effectiveness, we remove the upper circuit of rainy to rainy and the lower circuit of rain-free to rain-free, which are denoted as ``w/o Up'' and ``w/o Low'' in Table \ref{loss_ab}. The result indicates that both cycle-consistent circuits can enhance clean exemplars and offer supplementary constraints for improved rain removal.

\textbf{Effect of Different Scale Settings.} 
The multi-scale learning strategy provides the capability of scale-robust rain removal. We perform an ablation analysis of different scale settings as shown in Table \ref{scale}. It is observed that the combination of scales $r\in \{1,1/2,1/4\}$ yields the best results.


\begin{table}[!t]
\centering
\footnotesize
\caption{Ablation study for the choice of noise estimator on the real-world RDS test set.}
\label{ab_ddpm1}
\setlength{\tabcolsep}{0.8mm}{
\begin{tabular}{cccc}
\toprule
Estimator       & ResNet-50&VGG-16 &U-Net  \\ \midrule
PSNR / SSIM &21.07/0.629     &21.64/0.673     &\textbf{22.18}/\textbf{0.681}  \\
\bottomrule
\end{tabular}
}
\end{table}

\textbf{Choice of Different Noise Estimators.} 
Apart from U-Net \cite{ronneberger2015u}, we adopt other baselines as our scale noise estimators in DCDM, such as ResNet-50 \cite{he2016deep} and VGG-16 \cite{simonyan2014very}. Notably, all the kernels' parameters of ResNet-50 \cite{he2016deep} and VGG-16 \cite{simonyan2014very} are generated by our DG-Hyper. Table \ref{ab_ddpm1} indicates that U-Net is a fitting candidate for noise estimation.

\section{Limitations}
Although RainDiff shows superiority in five different rain degradations, it is unclear how its performance in diverse weather conditions, such as snow, low light, etc. In addition, similar to existing diffusion models \cite{ozdenizci2022restoring,saharia2022image}, it requires comparably longer runtime compared with end-to-end image restoration models which only require a single forward pass for processing without requisite steps involved during sampling. 
The time efficiency relies on the choice of algorithm hyper-parameters (e.g., a higher value of sampling steps increases image quality but also the inference time).


\section{Conclusion}
In this paper, we propose a new unpaired learning paradigm based on the diffusion model, called RainDiff, to tackle the unfavorable prevailing problem of real-world image deraining. The core of our method is a non-adversarial unpaired cycle-consistent architecture that can be trained using only unpaired data. Furthermore, we propose a degradation-conditioned diffusion model that learns multiple rain degradations for the diffusive generative process to improve the performance in image deraining for multiple degradations.
Experiments on both synthetic and real-world rainy images demonstrate the superiority of the proposed framework.

{
\bibliographystyle{splncs04}
\bibliography{main}

\begin{thebibliography}{10}
\providecommand{\url}[1]{\texttt{#1}}
\providecommand{\urlprefix}{URL }
\providecommand{\doi}[1]{https://doi.org/#1}

\bibitem{amit2021segdiff}
Amit, T., Nachmani, E., Shaharbany, T., Wolf, L.: Segdiff: Image segmentation with diffusion probabilistic models. arXiv preprint arXiv:2112.00390  (2021)

\bibitem{ba2022not}
Ba, Y., Zhang, H., Yang, E., Suzuki, A., Pfahnl, A., Chandrappa, C.C., de~Melo, C.M., You, S., Soatto, S., Wong, A., et~al.: Not just streaks: Towards ground truth for single image deraining. In: European Conference on Computer Vision. pp. 723--740. Springer (2022)

\bibitem{chang2017transformed}
Chang, Y., Yan, L., Zhong, S.: Transformed low-rank model for line pattern noise removal. In: Proceedings of the IEEE international conference on computer vision. pp. 1726--1734 (2017)

\bibitem{chen2016attention}
Chen, L.C., Yang, Y., Wang, J., Xu, W., Yuille, A.L.: Attention to scale: Scale-aware semantic image segmentation. In: Proceedings of the IEEE conference on computer vision and pattern recognition. pp. 3640--3649 (2016)

\bibitem{chen2022unpaired}
Chen, X., Pan, J., Jiang, K., Li, Y., Huang, Y., Kong, C., Dai, L., Fan, Z.: Unpaired deep image deraining using dual contrastive learning. In: Proceedings of the IEEE/CVF Conference on Computer Vision and Pattern Recognition. pp. 2017--2026 (2022)

\bibitem{chen2013generalized}
Chen, Y.L., Hsu, C.T.: A generalized low-rank appearance model for spatio-temporally correlated rain streaks. In: Proceedings of the IEEE International Conference on Computer Vision. pp. 1968--1975 (2013)

\bibitem{deng2020detail}
Deng, S., Wei, M., Wang, J., Feng, Y., Liang, L., Xie, H., Wang, F.L., Wang, M.: Detail-recovery image deraining via context aggregation networks. In: Proceedings of the IEEE/CVF conference on computer vision and pattern recognition. pp. 14560--14569 (2020)

\bibitem{fan2018residual}
Fan, Z., Wu, H., Fu, X., Huang, Y., Ding, X.: Residual-guide network for single image deraining. In: Proceedings of the 26th ACM international conference on Multimedia. pp. 1751--1759 (2018)

\bibitem{fu2017clearing}
Fu, X., Huang, J., Ding, X., Liao, Y., Paisley, J.: Clearing the skies: A deep network architecture for single-image rain removal. IEEE Transactions on Image Processing  \textbf{26}(6),  2944--2956 (2017)

\bibitem{detail_layer}
Fu, X., Huang, J., Zeng, D., Huang, Y., Ding, X., Paisley, J.W.: Removing rain from single images via a deep detail network. In: 2017 {IEEE} Conference on Computer Vision and Pattern Recognition. pp. 1715--1723 (2017)

\bibitem{fu2021rain}
Fu, X., Qi, Q., Zha, Z.J., Zhu, Y., Ding, X.: Rain streak removal via dual graph convolutional network. In: Proc. AAAI Conf. Artif. Intell. pp.~1--9 (2021)

\bibitem{gu2017joint}
Gu, S., Meng, D., Zuo, W., Zhang, L.: Joint convolutional analysis and synthesis sparse representation for single image layer separation. In: Proceedings of the IEEE International Conference on Computer Vision. pp. 1708--1716 (2017)

\bibitem{gu2022vector}
Gu, S., Chen, D., Bao, J., Wen, F., Zhang, B., Chen, D., Yuan, L., Guo, B.: Vector quantized diffusion model for text-to-image synthesis. In: Proceedings of the IEEE/CVF Conference on Computer Vision and Pattern Recognition. pp. 10696--10706 (2022)

\bibitem{45803}
Ha, D., Dai, A., Le, Q.: Hypernetworks (2016)

\bibitem{hadsell2006dimensionality}
Hadsell, R., Chopra, S., LeCun, Y.: Dimensionality reduction by learning an invariant mapping. In: 2006 IEEE computer society conference on computer vision and pattern recognition (CVPR'06). vol.~2, pp. 1735--1742. IEEE (2006)

\bibitem{han2020decomposed}
Han, K., Xiang, X.: Decomposed cyclegan for single image deraining with unpaired data. In: ICASSP 2020-2020 IEEE International Conference on Acoustics, Speech and Signal Processing (ICASSP). pp. 1828--1832. IEEE (2020)

\bibitem{he2020momentum}
He, K., Fan, H., Wu, Y., Xie, S., Girshick, R.: Momentum contrast for unsupervised visual representation learning. In: Proceedings of the IEEE/CVF conference on computer vision and pattern recognition. pp. 9729--9738 (2020)

\bibitem{he2016deep}
He, K., Zhang, X., Ren, S., Sun, J.: Deep residual learning for image recognition. In: Proceedings of the IEEE conference on computer vision and pattern recognition. pp. 770--778 (2016)

\bibitem{ho2020denoising}
Ho, J., Jain, A., Abbeel, P.: Denoising diffusion probabilistic models. Advances in Neural Information Processing Systems  \textbf{33},  6840--6851 (2020)

\bibitem{hu2019depth}
Hu, X., Fu, C.W., Zhu, L., Heng, P.A.: Depth-attentional features for single-image rain removal. In: Proceedings of the IEEE/CVF Conference on computer vision and pattern recognition. pp. 8022--8031 (2019)

\bibitem{huynh2008scope}
Huynh-Thu, Q., Ghanbari, M.: Scope of validity of psnr in image/video quality assessment. Electronics letters  \textbf{44}(13),  800--801 (2008)

\bibitem{jin2019unsupervised}
Jin, X., Chen, Z., Lin, J., Chen, Z., Zhou, W.: Unsupervised single image deraining with self-supervised constraints. In: 2019 IEEE International Conference on Image Processing (ICIP). pp. 2761--2765. IEEE (2019)

\bibitem{li2018benchmarking}
Li, B., Ren, W., Fu, D., Tao, D., Feng, D., Zeng, W., Wang, Z.: Benchmarking single-image dehazing and beyond. IEEE Transactions on Image Processing  \textbf{28}(1),  492--505 (2018)

\bibitem{li2022all}
Li, B., Liu, X., Hu, P., Wu, Z., Lv, J., Peng, X.: All-in-one image restoration for unknown corruption. In: Proceedings of the IEEE/CVF Conference on Computer Vision and Pattern Recognition. pp. 17452--17462 (2022)

\bibitem{li2018non}
Li, G., He, X., Zhang, W., Chang, H., Dong, L., Lin, L.: Non-locally enhanced encoder-decoder network for single image de-raining. In: Proceedings of the 26th ACM international conference on Multimedia. pp. 1056--1064 (2018)

\bibitem{li2018recurrent}
Li, X., Wu, J., Lin, Z., Liu, H., Zha, H.: Recurrent squeeze-and-excitation context aggregation net for single image deraining. In: Proceedings of the European Conference on Computer Vision (ECCV). pp. 254--269 (2018)

\bibitem{li2016rain}
Li, Y., Tan, R.T., Guo, X., Lu, J., Brown, M.S.: Rain streak removal using layer priors. In: Proceedings of the IEEE conference on computer vision and pattern recognition. pp. 2736--2744 (2016)

\bibitem{liu2020deep}
Liu, L., Ouyang, W., Wang, X., Fieguth, P., Chen, J., Liu, X., Pietik{\"a}inen, M.: Deep learning for generic object detection: A survey. International journal of computer vision  \textbf{128}(2),  261--318 (2020)

\bibitem{luo2015removing}
Luo, Y., Xu, Y., Ji, H.: Removing rain from a single image via discriminative sparse coding. In: Proceedings of the IEEE international conference on computer vision. pp. 3397--3405 (2015)

\bibitem{nichol2022glide}
Nichol, A.Q., Dhariwal, P., Ramesh, A., Shyam, P., Mishkin, P., Mcgrew, B., Sutskever, I., Chen, M.: Glide: Towards photorealistic image generation and editing with text-guided diffusion models. In: International Conference on Machine Learning. pp. 16784--16804. PMLR (2022)

\bibitem{ozdenizci2022restoring}
{\"O}zdenizci, O., Legenstein, R.: Restoring vision in adverse weather conditions with patch-based denoising diffusion models. arXiv preprint arXiv:2207.14626  (2022)

\bibitem{quan2021removing}
Quan, R., Yu, X., Liang, Y., Yang, Y.: Removing raindrops and rain streaks in one go. In: Proceedings of the IEEE/CVF Conference on Computer Vision and Pattern Recognition. pp. 9147--9156 (2021)

\bibitem{ren2019progressive}
Ren, D., Zuo, W., Hu, Q., Zhu, P., Meng, D.: Progressive image deraining networks: a better and simpler baseline. In: Proceedings of the IEEE Conference on Computer Vision and Pattern Recognition. pp. 3937--3946 (2019)

\bibitem{ronneberger2015u}
Ronneberger, O., Fischer, P., Brox, T.: U-net: Convolutional networks for biomedical image segmentation. In: International Conference on Medical image computing and computer-assisted intervention. pp. 234--241. Springer (2015)

\bibitem{saharia2022image}
Saharia, C., Ho, J., Chan, W., Salimans, T., Fleet, D.J., Norouzi, M.: Image super-resolution via iterative refinement. IEEE Transactions on Pattern Analysis and Machine Intelligence  (2022)

\bibitem{simonyan2014very}
Simonyan, K., Zisserman, A.: Very deep convolutional networks for large-scale image recognition. arXiv preprint arXiv:1409.1556  (2014)

\bibitem{sohl2015deep}
Sohl-Dickstein, J., Weiss, E., Maheswaranathan, N., Ganguli, S.: Deep unsupervised learning using nonequilibrium thermodynamics. In: International Conference on Machine Learning. pp. 2256--2265. PMLR (2015)

\bibitem{song2020denoising}
Song, J., Meng, C., Ermon, S.: Denoising diffusion implicit models. arXiv preprint arXiv:2010.02502  (2020)

\bibitem{song2019generative}
Song, Y., Ermon, S.: Generative modeling by estimating gradients of the data distribution. Advances in Neural Information Processing Systems  \textbf{32} (2019)

\bibitem{sultani2018real}
Sultani, W., Chen, C., Shah, M.: Real-world anomaly detection in surveillance videos. In: Proceedings of the IEEE conference on computer vision and pattern recognition. pp. 6479--6488 (2018)

\bibitem{wang2019erl}
Wang, G., Sun, C., Sowmya, A.: Erl-net: Entangled representation learning for single image de-raining. In: Proceedings of the IEEE International Conference on Computer Vision. pp. 5644--5652 (2019)

\bibitem{wang2019spatial}
Wang, T., Yang, X., Xu, K., Chen, S., Zhang, Q., Lau, R.W.: Spatial attentive single-image deraining with a high quality real rain dataset. In: Proceedings of the IEEE Conference on Computer Vision and Pattern Recognition. pp. 12270--12279 (2019)

\bibitem{wang2017hierarchical}
Wang, Y., Liu, S., Chen, C., Zeng, B.: A hierarchical approach for rain or snow removing in a single color image. IEEE Transactions on Image Processing  \textbf{26}(8),  3936--3950 (2017)

\bibitem{wang2004image}
Wang, Z., Bovik, A.C., Sheikh, H.R., Simoncelli, E.P.: Image quality assessment: from error visibility to structural similarity. IEEE transactions on image processing  \textbf{13}(4),  600--612 (2004)

\bibitem{wei2019semi}
Wei, W., Meng, D., Zhao, Q., Xu, Z., Wu, Y.: Semi-supervised transfer learning for image rain removal. In: Proceedings of the IEEE Conference on Computer Vision and Pattern Recognition. pp. 3877--3886 (2019)

\bibitem{wei2021deraincyclegan}
Wei, Y., Zhang, Z., Wang, Y., Xu, M., Yang, Y., Yan, S., Wang, M.: Deraincyclegan: Rain attentive cyclegan for single image deraining and rainmaking. IEEE Transactions on Image Processing  \textbf{30},  4788--4801 (2021)

\bibitem{wei2021semi}
Wei, Y., Zhang, Z., Wang, Y., Zhang, H., Zhao, M., Xu, M., Wang, M.: Semi-deraingan: A new semi-supervised single image deraining. In: 2021 IEEE International Conference on Multimedia and Expo (ICME). pp.~1--6. IEEE (2021)

\bibitem{welling2011bayesian}
Welling, M., Teh, Y.W.: Bayesian learning via stochastic gradient langevin dynamics. In: Proceedings of the 28th international conference on machine learning (ICML-11). pp. 681--688 (2011)

\bibitem{wojna2019devil}
Wojna, Z., Ferrari, V., Guadarrama, S., Silberman, N., Chen, L.C., Fathi, A., Uijlings, J.: The devil is in the decoder: Classification, regression and gans. International Journal of Computer Vision  \textbf{127}(11),  1694--1706 (2019)

\bibitem{wu2021contrastive}
Wu, H., Qu, Y., Lin, S., Zhou, J., Qiao, R., Zhang, Z., Xie, Y., Ma, L.: Contrastive learning for compact single image dehazing. In: Proceedings of the IEEE/CVF Conference on Computer Vision and Pattern Recognition. pp. 10551--10560 (2021)

\bibitem{yang2017deep}
Yang, W., Tan, R.T., Feng, J., Liu, J., Guo, Z., Yan, S.: Deep joint rain detection and removal from a single image. In: Proceedings of the IEEE Conference on Computer Vision and Pattern Recognition. pp. 1357--1366 (2017)

\bibitem{yasarla2020syn2real}
Yasarla, R., Sindagi, V.A., Patel, V.M.: Syn2real transfer learning for image deraining using gaussian processes. In: Proceedings of the IEEE/CVF conference on computer vision and pattern recognition. pp. 2726--2736 (2020)

\bibitem{ye2021closing}
Ye, Y., Chang, Y., Zhou, H., Yan, L.: Closing the loop: Joint rain generation and removal via disentangled image translation. In: Proceedings of the IEEE/CVF Conference on Computer Vision and Pattern Recognition. pp. 2053--2062 (2021)

\bibitem{ye2022unsupervised}
Ye, Y., Yu, C., Chang, Y., Zhu, L., Zhao, X.L., Yan, L., Tian, Y.: Unsupervised deraining: Where contrastive learning meets self-similarity. In: Proceedings of the IEEE/CVF Conference on Computer Vision and Pattern Recognition. pp. 5821--5830 (2022)

\bibitem{yu2017dilated}
Yu, F., Koltun, V., Funkhouser, T.: Dilated residual networks. In: Proceedings of the IEEE conference on computer vision and pattern recognition. pp. 472--480 (2017)

\bibitem{zhang2018density}
Zhang, H., Patel, V.M.: Density-aware single image de-raining using a multi-stream dense network. In: Proceedings of the IEEE conference on computer vision and pattern recognition. pp. 695--704 (2018)

\bibitem{zhang2019image}
Zhang, H., Sindagi, V., Patel, V.M.: Image de-raining using a conditional generative adversarial network. IEEE transactions on circuits and systems for video technology  \textbf{30}(11),  3943--3956 (2019)

\bibitem{zhu2019singe}
Zhu, H., Peng, X., Zhou, J.T., Yang, S., Chanderasekh, V., Li, L., Lim, J.H.: Singe image rain removal with unpaired information: A differentiable programming perspective. In: Proceedings of the AAAI Conference on Artificial Intelligence. vol.~33, pp. 9332--9339 (2019)

\bibitem{zhu2017unpaired}
Zhu, J.Y., Park, T., Isola, P., Efros, A.A.: Unpaired image-to-image translation using cycle-consistent adversarial networks. In: Proceedings of the IEEE international conference on computer vision. pp. 2223--2232 (2017)

\end{thebibliography}
}
\end{document}


\title{Rethinking Real-world Image Deraining via An Unpaired Degradation-Conditioned Diffusion Model \\ Supplementary Material} 

\titlerunning{Abbreviated paper title}

\author{First Author\inst{1}\orcidlink{0000-1111-2222-3333} \and
Second Author\inst{2,3}\orcidlink{1111-2222-3333-4444} \and
Third Author\inst{3}\orcidlink{2222--3333-4444-5555}}

\authorrunning{F.~Author et al.}

\institute{Princeton University, Princeton NJ 08544, USA \and
Springer Heidelberg, Tiergartenstr.~17, 69121 Heidelberg, Germany
\email{lncs@springer.com}\\
\url{http://www.springer.com/gp/computer-science/lncs} \and
ABC Institute, Rupert-Karls-University Heidelberg, Heidelberg, Germany\\
\email{\{abc,lncs\}@uni-heidelberg.de}}

\maketitle

This supplementary material contains ten parts:
\begin{enumerate}
  \item Discussions with Closely-Related Methods
  \item Applications
  \item Balanced Weight Analysis
  \item Choice of Different Non-diffusive Generators
  \item Effect of Different Rain Degradations
  \item Failure Cases
  \item Additional Qualitative Results
  \item Extending to Other Image Translation Tasks
  \item Comparison with More Conditioning Methods
  \item Analysis for Stability
\end{enumerate}

\section{Discussions with Closely-Related Methods}
To better solve real-world image deraining problems, RainDiff adopts an effective unpaired cycle-consistent architecture with a degradation-conditioned diffusion model. In this Section, we will compare and engage in in-depth discussions with the most closely-related approaches.

In unpaired learning, the circulatory structures with cycle-consistency loss functions are commonly used for model training \cite{chen2022unpaired,wei2021deraincyclegan,zhu2017unpaired}. Different from existing popular circulatory structures, our method requires no discriminators for adversarial training and provides a reliable non-adversarial training process.
Additionally, the generators $G_{\phi}^{A}$ and $G_{\phi}^{B}$ are both only used in the training phase and remain uninvolved in the testing phase for image deraining. Furthermore, our method can well generalize to real-world rainy images as exhibited in Figs. \ref{fig:real_rd}-\ref{fig:real_google}.

\begin{table}[!h]
\centering
\footnotesize
\caption{Quantitative results of different diffusion models on the real-world test sets.}
\label{ab_ddpm}
\setlength{\tabcolsep}{0.4mm}{
\begin{tabular}{c|c|c|c|c}
\toprule
\multirow{2}{*}{Model} & RS                                                                                  & RD                                                                                  & RDS                                                                                 & RHS                                                                                 \\
                       & PSNR/SSIM                                                                           & PSNR/SSIM                                                                           & PSNR/SSIM                                                                           & PSNR/SSIM                                                                           \\ \midrule
SR3                    & 22.46/0.716                                                                         & 19.32/0.603                                                                         & 20.24/0.642                                                                         & 20.01/0.619                                                                         \\
WeatherDiffusion       & 22.08/0.721                                                                         & 19.55/0.617                                                                         & 20.69/0.648                                                                         & 20.28/0.623                                                                         \\
Ours                   & \textbf{24.73}/\textbf{0.770} & \textbf{21.89}/\textbf{0.685} & \textbf{22.18}/\textbf{0.681} & \textbf{22.66}/\textbf{0.694} \\ \bottomrule
\end{tabular}
\end{table}

We note that a handful of diffusion models for image restoration have recently emerged. To validate the superiority of the proposed degradation-conditioned diffusion model (DCDM), we choose two recent diffusion-based image restoration methods, i.e., SR3 \cite{saharia2022image} and WeatherDiffusion \cite{ozdenizci2022restoring}, as the competitors. For fair comparisons, we substitute DCDM with these two diffusion models and retrain the entire paradigm on the mixture dataset. Table \ref{ab_ddpm} demonstrates the superiority of the proposed method. It affirms the significance of multi-degradation/scale rain information captured by DCDM.

\begin{figure*}[!t]
	\centering
\footnotesize
	\subfigure[]{
		\includegraphics[width=1.5in, height=1.2in]{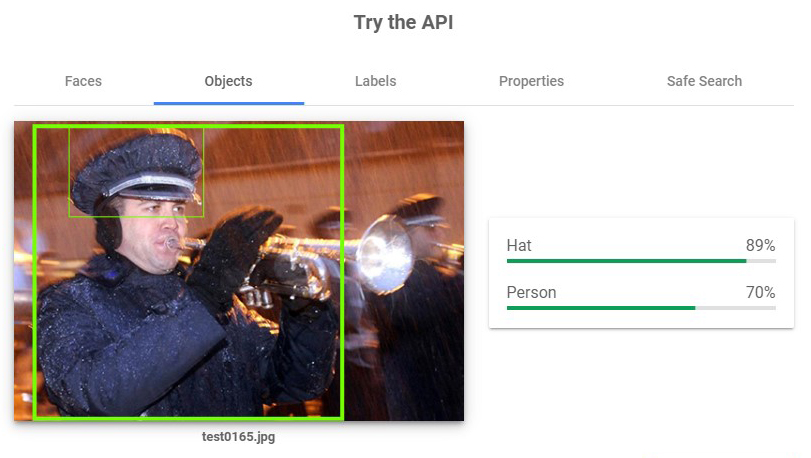}
	}
	\subfigure[]{
		\includegraphics[width=1.5in, height=1.2in]{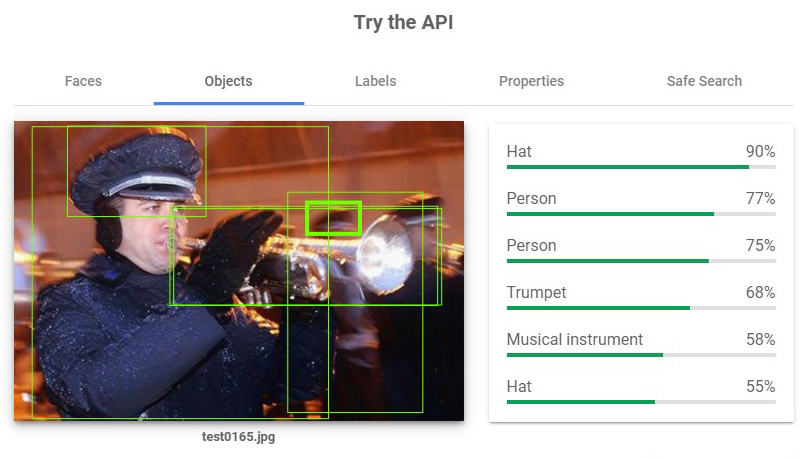}
	}
	\subfigure[]{
		\includegraphics[width=1.5in, height=1.2in]{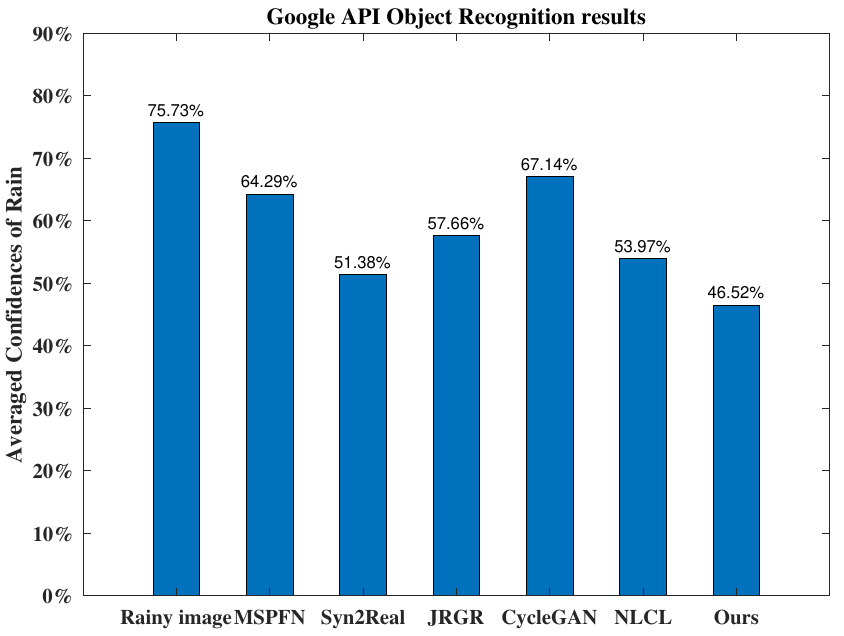}
	}
	\caption{Deraining results tested on Google Vision API. From (a)-(c): (a) the object recognition result in the real-world rainy image, (b) the object recognition result after deraining by our RainDiff, and (c) the averaged confidences in recognizing rain from 400 sets of the real-world rainy images and derained images of MSPFN \cite{jiang2020multi}, Syn2Real \cite{yasarla2020syn2real}, JRGR \cite{ye2021closing}, CycleGAN\cite{zhu2017unpaired}, NLCL\cite{ye2022unsupervised} and our RainDiff respectively.}
	\label{fig:google_api}
\end{figure*}
\section{Applications} 
To demonstrate that our RainDiff can benefit vision-based applications, we employ Google Vision API to evaluate the deraining results. One of the results is shown in Fig. \ref{fig:google_api} (a-b) where Google Vision API can recognize the rainy weather in the rainy image while it cannot recognize the rainy weather in the deraining image. Furthermore, we use Google Vision API to test 400 sets of real-world rainy images and derained images of MSPFN \cite{jiang2020multi}, Syn2Real \cite{yasarla2020syn2real}, JRGR \cite{ye2021closing}, CycleGAN\cite{zhu2017unpaired}, NLCL \cite{ye2022unsupervised} and our RainDiff in Fig. \ref{fig:google_api} (c). After deraining, the confidence in recognizing rain from the images is significantly reduced.

\begin{figure}[htbp] \centering
	\includegraphics[width=0.8\linewidth]{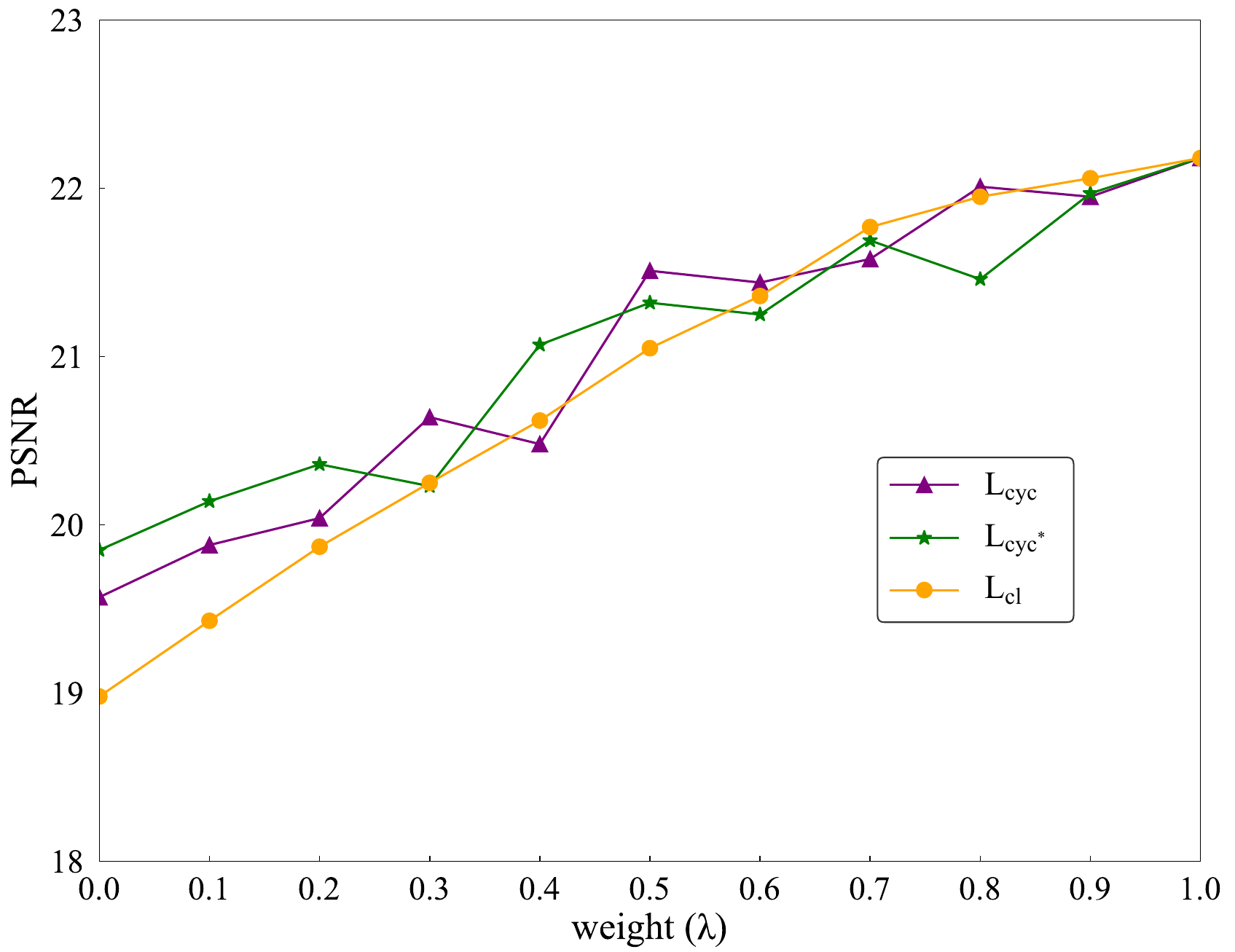}
	\caption{Effectiveness of different balanced weights.}
	\label{fig:weight_cyc}
\end{figure}

\section{Balanced Weight Analysis}
In this section, we evaluate the different settings of the balanced weights $\lambda_{cyc}$, $\lambda_{cyc^{*}}$ and $\lambda_{cl}$ on the real-world RDS test set. As demonstrated in Fig. \ref{fig:weight_cyc}, all the balanced weights are chosen experimentally, which effectively optimizes our network.



\begin{table}[!h]
 \renewcommand\arraystretch{1.0}
\centering
\caption{Ablation study for the choice of non-diffusive generators on the real-world test sets.}
\label{com_ng}
\begin{tabular}{c|c|c|c|c}
\toprule
\multirow{2}{*}{Model} & RS                                                                                  & RD                                                                                  & RDS                                                                                 & RHS                                                                                 \\
                       & PSNR/SSIM                                                                           & PSNR/SSIM                                                                           & PSNR/SSIM                                                                           & PSNR/SSIM                                                                           \\ \midrule
VGG-19                   & 22.85/0.707                                                                         & 20.52/0.619                                                                         & 20.86/0.631                                                                         & 20.25/0.628                                                                        \\
Transformer      & 23.97/0.736                                                                        & 21.04/0.630                                                                         & 21.56/0.659                                                                        & 21.81/0.645                                                                         \\               
Ours                   & \textbf{24.73}/\textbf{0.770} & \textbf{21.89}/\textbf{0.685} & \textbf{22.18}/\textbf{0.681} & \textbf{22.66}/\textbf{0.694} \\ \bottomrule
\end{tabular}
\end{table}
\section{Choice of Different Non-diffusive Generators}
We use U-Net as non-diffusive generators due to its effectiveness proved by some cycle-consistent methods (e.g., JRGR and DerainCycleGAN). 
We added an ablation study using alternative backbones (i.e., Transformer and VGG-19) as shown in Table \ref{com_ng}.

\begin{table}[!t]
\centering
\footnotesize
\caption{Ablation analysis for different rain degradations on the real-world test sets. w/o Real refers to without all real-world sets.}
\label{ab_rds}
\setlength{\tabcolsep}{1.2mm}{
\begin{tabular}{ccccc}
\toprule
\multirow{2}{*}{Model} & RS                                                              & RD                                                              & RDS                                                             & RHS                                                             \\
                       & PSNR/SSIM                                                       & PSNR/SSIM                                                       & PSNR/SSIM                                                       & PSNR/SSIM                                                       \\ \midrule
w/o RS                 & 22.36/0.726                                                     & \textbf{21.96}/\textbf{0.689}                                                     & 21.48/0.663                                                     & 20.92/0.645                                                     \\
w/o RH                 & \textbf{24.80}/0.774                                                     & 21.94/0.682                                                     & \textbf{22.21}/\textbf{0.684}                                                     & 18.07/0.588                                                     \\
w/o RD                 & 24.78/\textbf{0.771}                                                     & 18.29/0.554                                                     & 18.96/0.590                                                     & 22.64/\textbf{0.695}                                                   \\
w/o RHS                & 23.65/0.742                                                     & 21.83/0.682                                                     & 21.87/0.673                                                     & 19.39/0.627                                                     \\
w/o RDS                & 23.89/0.750                                                     & 21.05/0.676                                                     & 18.11/0.603                                                     & 21.49/0.662                                                     \\
w/o Real               & 23.86/0.747                                                     & 20.57/0.663                                                     & 20.25/0.654                                                     & 20.61/0.643                                                     \\
Ours                   & 24.73/0.770 & 21.89/0.685 & 22.18/0.681 & \textbf{22.66}/0.694 \\ \bottomrule
\end{tabular}}
\end{table}

\section{Effect of Different Rain Degradations}
In this section, we carry out experiments to demonstrate
the effectiveness of RainDiff on different rain degradations. As shown in Table \ref{ab_rds}, we make the following observations: 1) If there is a significant domain gap between two different rain degradations, excluding one of them during training can enhance the deraining performance of the other. For instance, we exclude RS train sets during training to achieve improved performance on the RD test set. This marginal improvement serves as evidence that our DCDM effectively addresses diverse rain degradations. 2) If the mixture of rain train sets (i.e., RDS and RHS) is removed, the deraining performance on all types of rain degradations will decrease. This implies that the combination of different rain characteristics can provide additional benefits for removing specific types of rain degradation. 3) The removal of real-world train sets leads to a noticeable decrease in performance. RainDiff leverages unpaired real-world rain data during training to enhance the deraining performance under realistic rainy conditions.

\begin{figure*}[!t]
	\centering
	\includegraphics[width=1.0\linewidth]{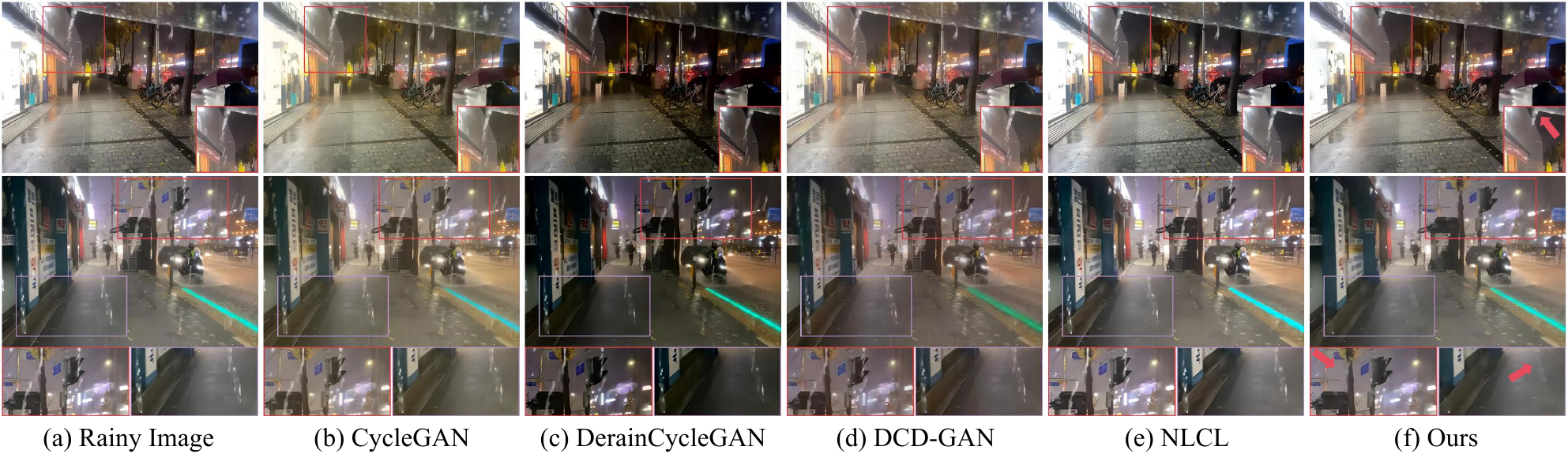}
	\caption{Typical failure cases of different methods. The red arrows point to some traces of raindrops that are not removed.} 
	\label{fig:real_fail}
\end{figure*}

\section{Failure Cases}
Fig. \ref{fig:real_fail} exhibits some typical failure cases of different algorithms. It is observed that our method may fail to remove some unseen rain degradations, such as these dripping raindrops. Despite these limitations, our approach still outperforms state-of-the-art (SOTA) approaches and is capable of removing the majority of raindrops. But there may remain some traces of raindrops that are not removed.
One possible solution for these failure cases may incorporate more different rain degradations for training, to achieve better rain removal performance.


\section{Additional Qualitative Results} 
\begin{figure*}[!t]
	\centering
	\includegraphics[width=1.0\linewidth]{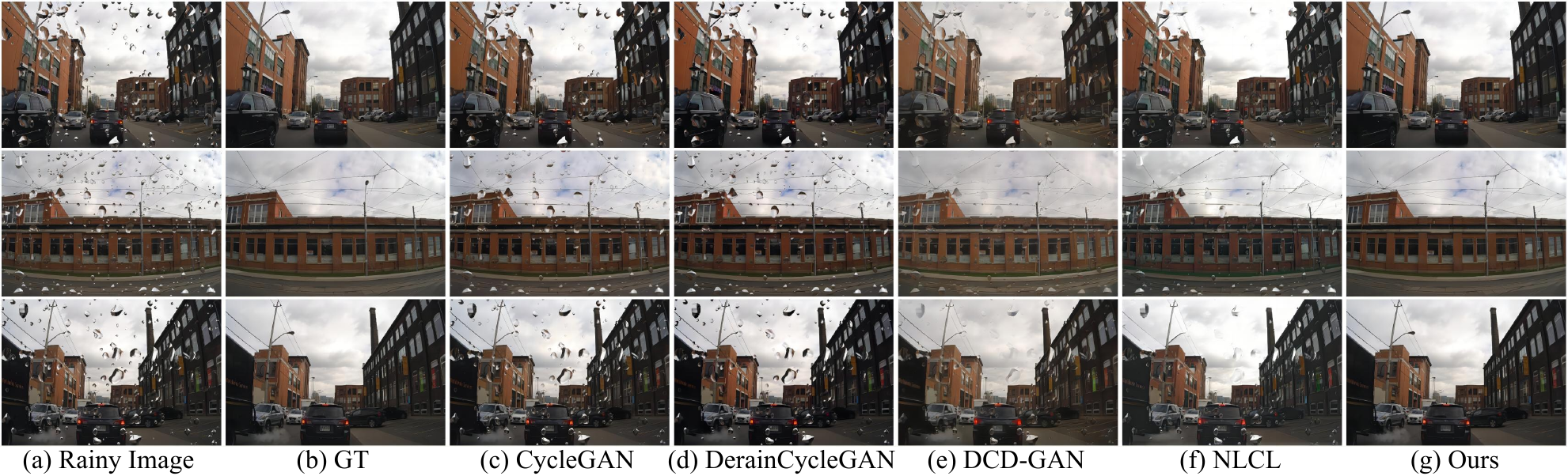}
	\caption{Visual comparison results on the synthetic RD test set.} 
	\label{fig:syn_rd}
\end{figure*}

\begin{figure*}[!t]
	\centering
	\includegraphics[width=1.0\linewidth]{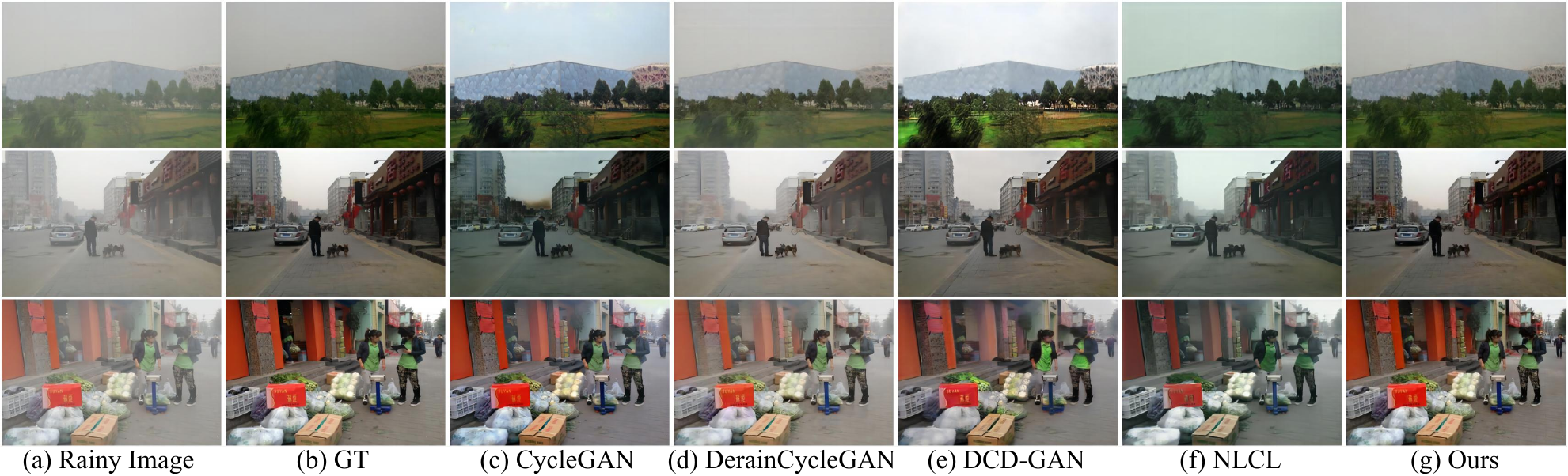}
	\caption{Visual comparison results on the synthetic RH test set.} 
	\label{fig:syn_rh}
\end{figure*}

\begin{figure*}[!t]
	\centering
	\includegraphics[width=1.0\linewidth]{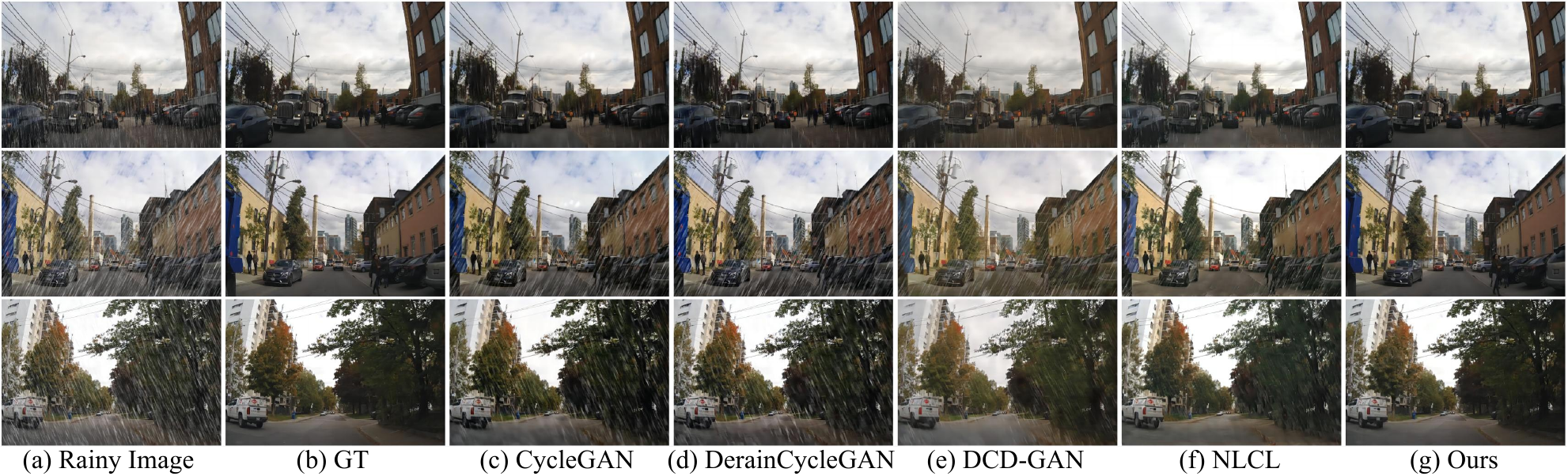}
	\caption{Visual comparison results on the synthetic RS test set.} 
	\label{fig:syn_rs}
\end{figure*}

\begin{figure*}[!t]
	\centering
	\includegraphics[width=1.0\linewidth]{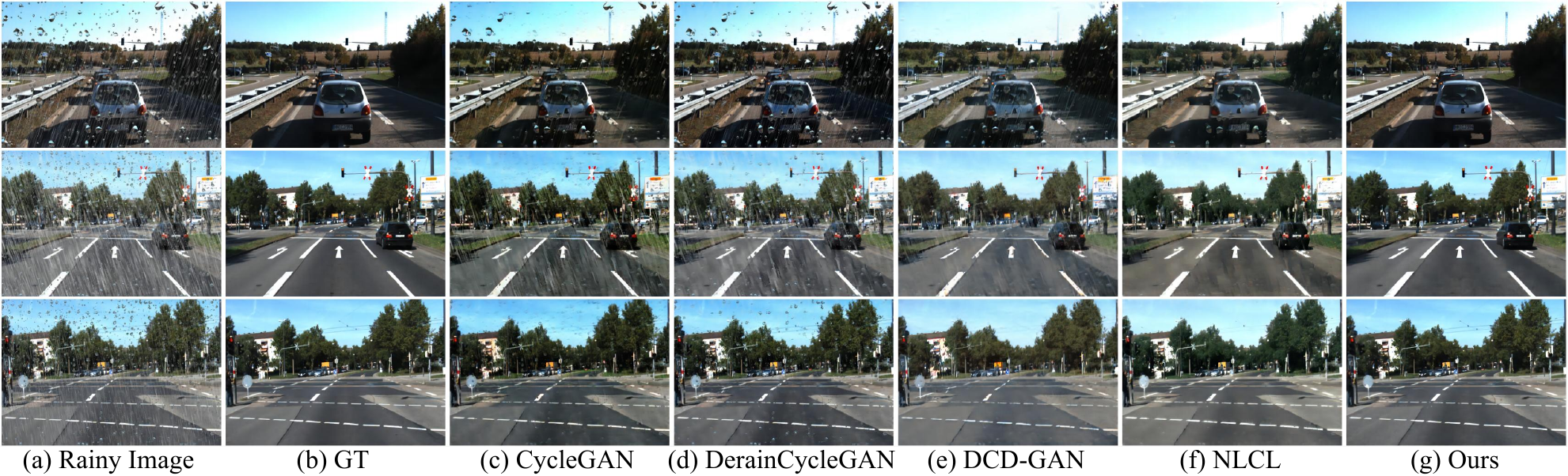}
	\caption{Visual comparison results on the synthetic RDS test set.} 
	\label{fig:syn_rds}
\end{figure*}

\begin{figure*}[!t]
	\centering
	\includegraphics[width=1.0\linewidth]{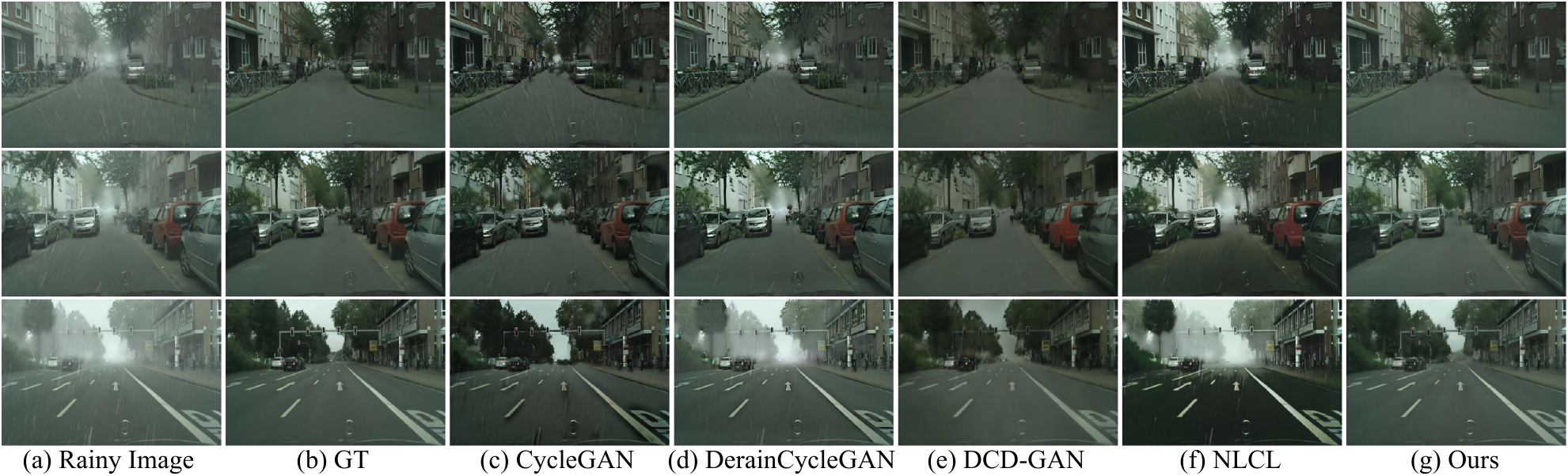}
	\caption{Visual comparison results on the synthetic RSH test set.} 
	\label{fig:syn_rsh}
\end{figure*}

Figs. \ref{fig:syn_rd}-\ref{fig:syn_rsh} illustrates the visual results of the five synthetic test sets. It is observed that existing unpaired deraining methods either fail to remove rain or generate blurry background images, while our method can successfully remove rain and produce visually pleasing rain-free images.

\begin{figure*}[!t]
	\centering
	\includegraphics[width=1.0\linewidth]{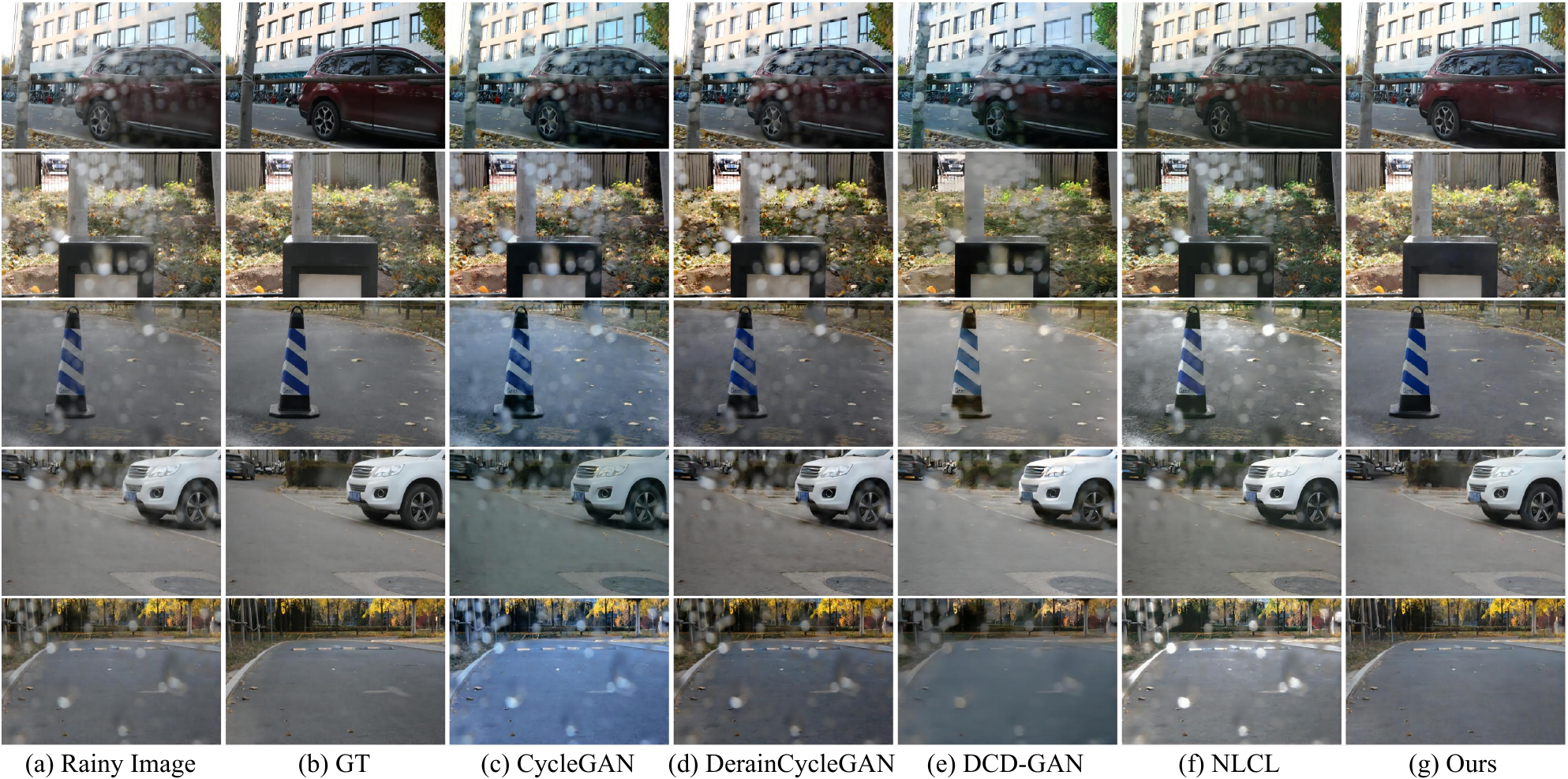}
	\caption{Visual comparison results on the real-world RD test set.} 
	\label{fig:real_rd}
\end{figure*}

\begin{figure*}[!t]
	\centering
	\includegraphics[width=1.0\linewidth]{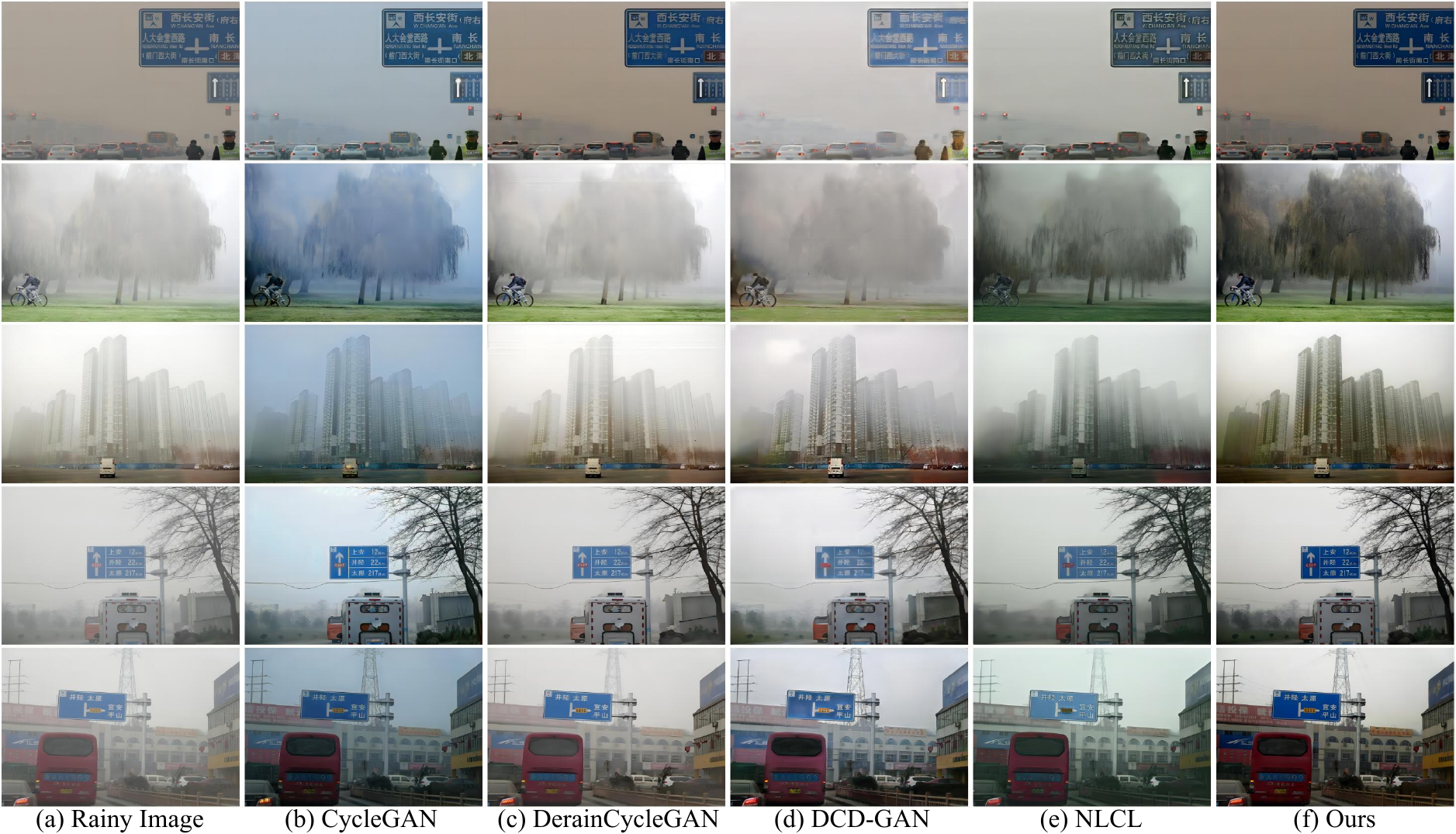}
	\caption{Visual comparison results on the real-world RH test set.} 
	\label{fig:real_rh}
\end{figure*}

\begin{figure*}[!t]
	\centering
	\includegraphics[width=1.0\linewidth]{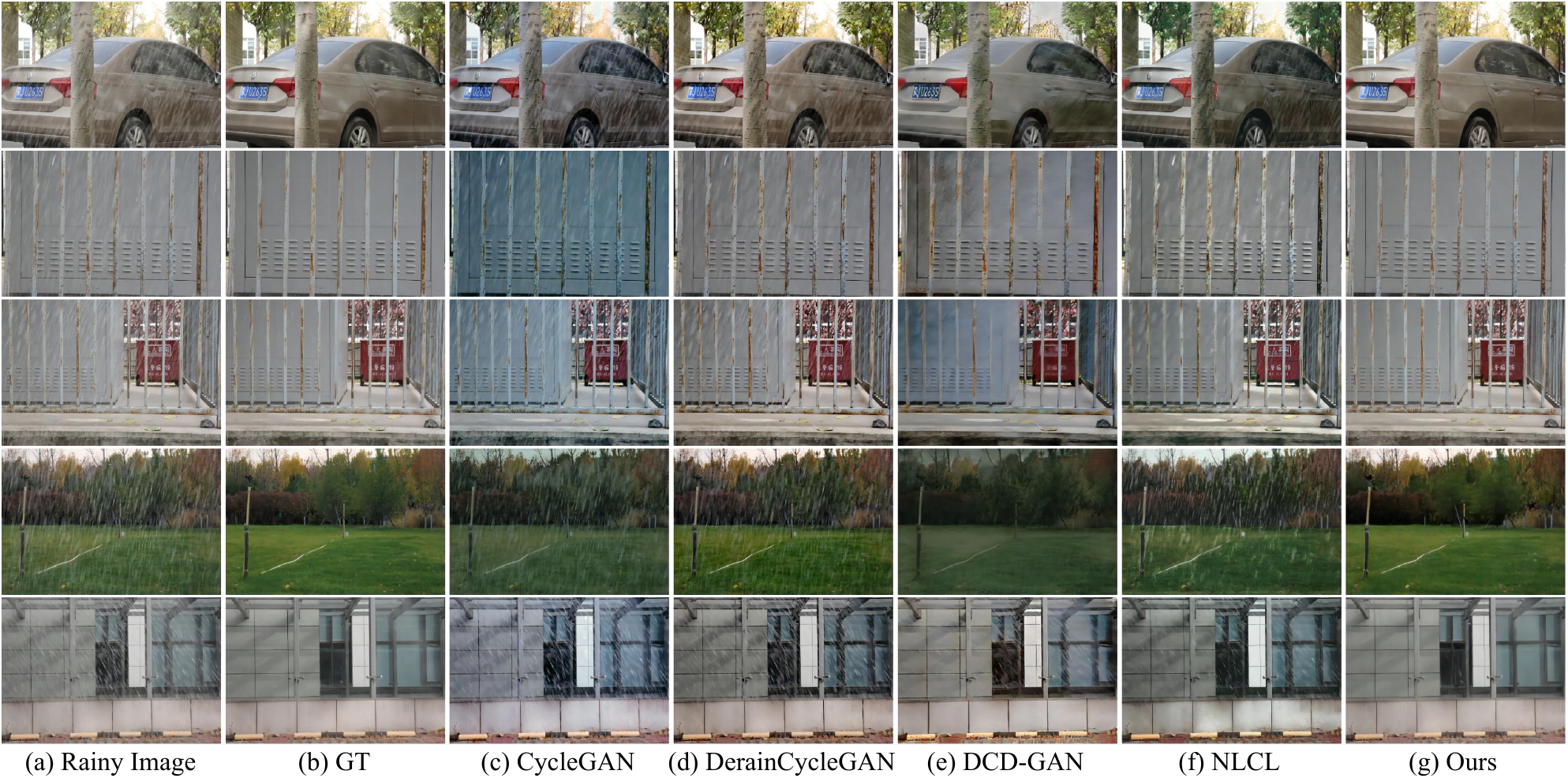}
	\caption{Visual comparison results on the real-world RS test set.} 
	\label{fig:real_rs}
\end{figure*}

\begin{figure*}[!t]
	\centering
	\includegraphics[width=1.0\linewidth]{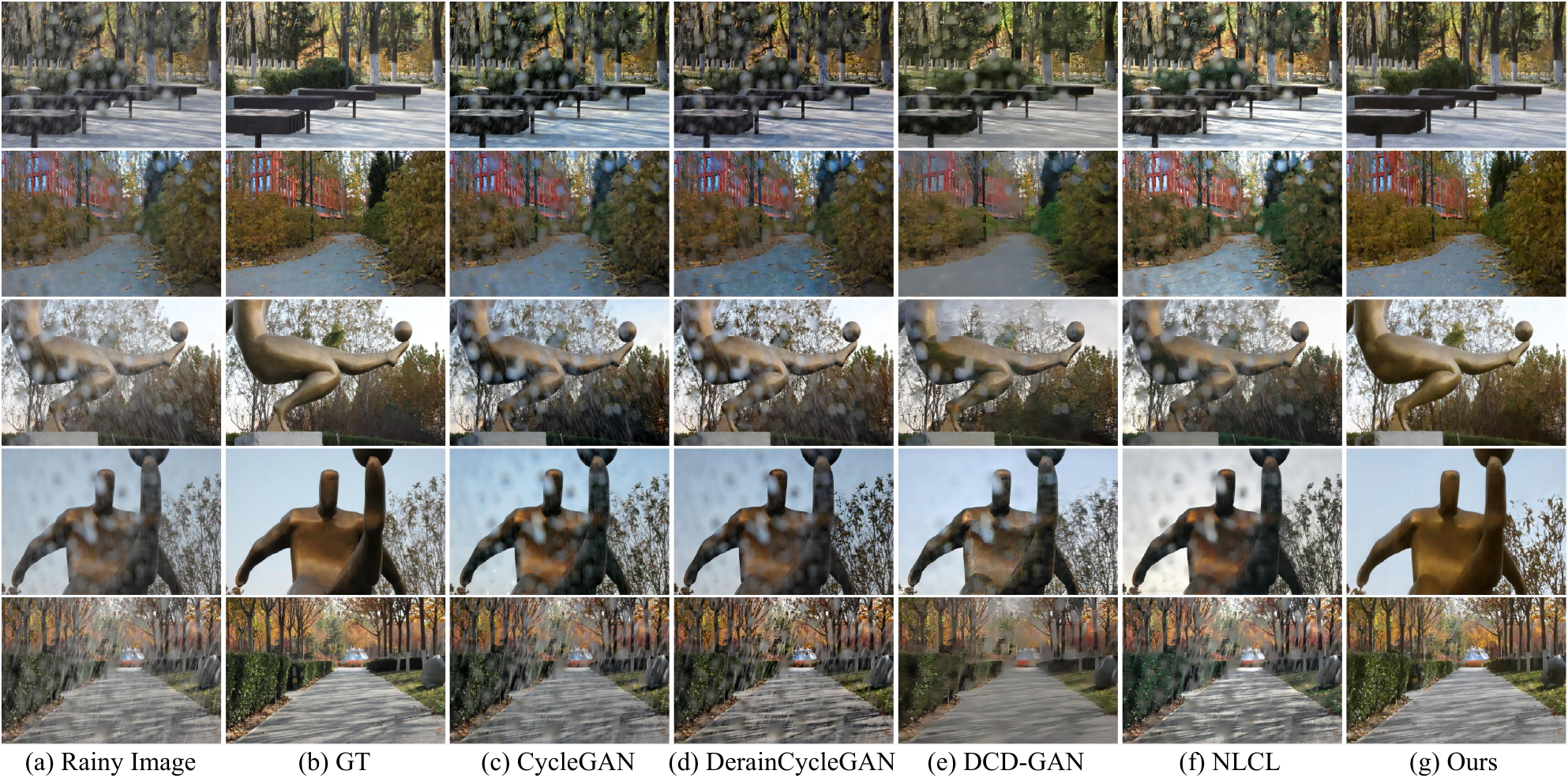}
	\caption{Visual comparison results on the real-world RDS test set.} 
	\label{fig:real_rds}
\end{figure*}

\begin{figure*}[!t]
	\centering
	\includegraphics[width=1.0\linewidth]{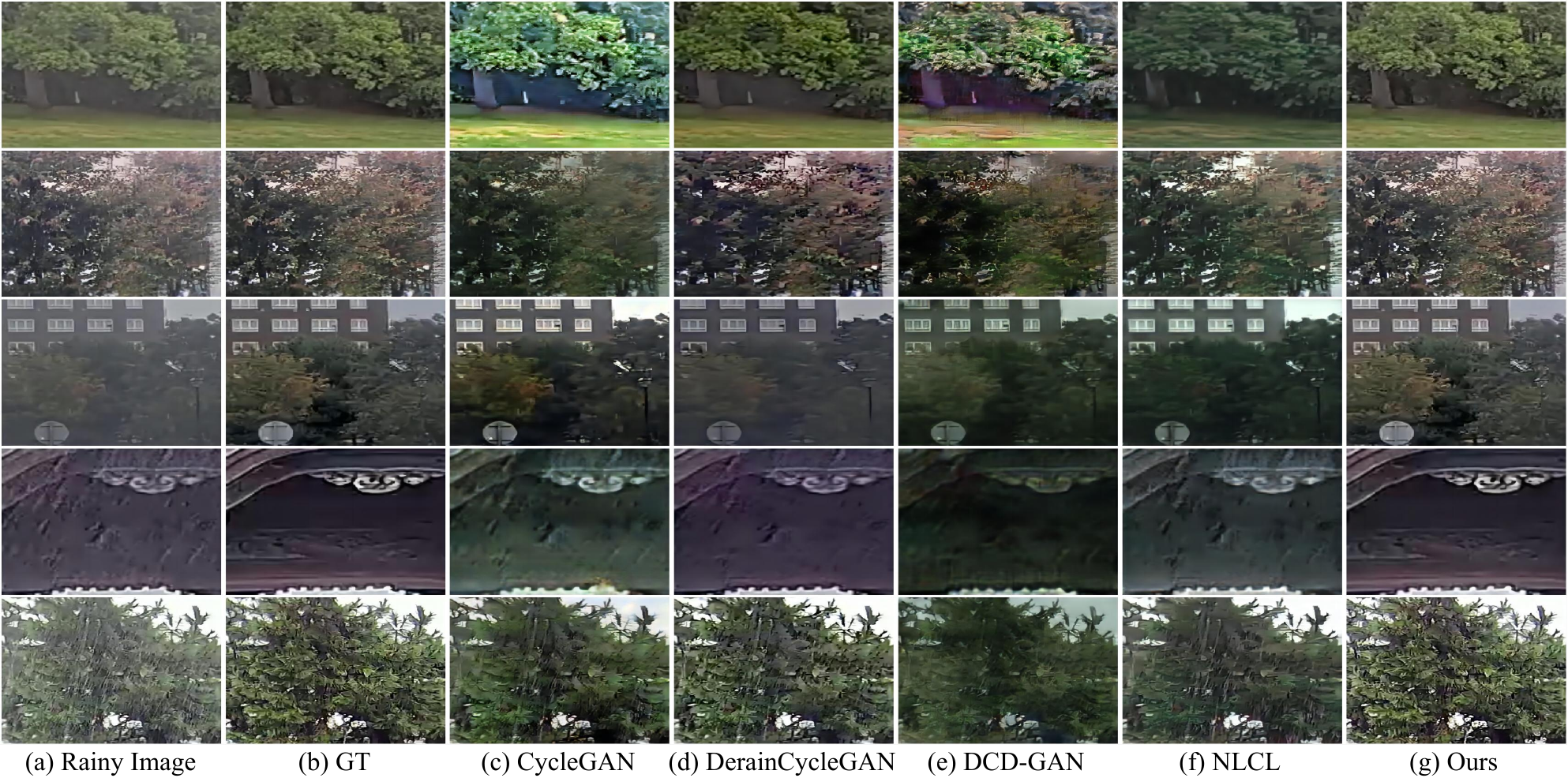}
	\caption{Visual comparison results on the real-world RSH test set.} 
	\label{fig:real_rsh}
\end{figure*}

\begin{figure*}[!t]
	\centering
	\includegraphics[width=1.0\linewidth]{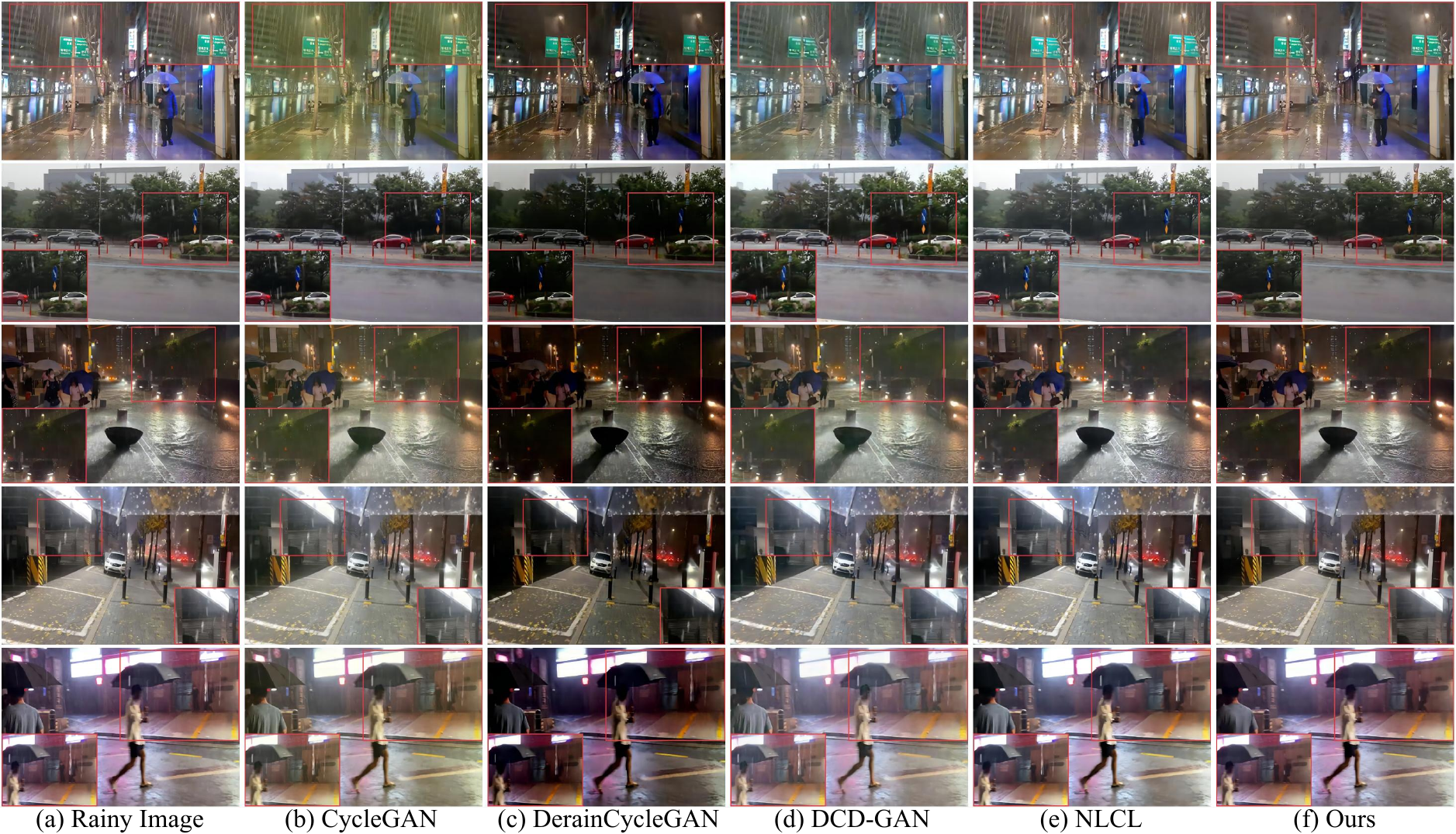}
	\caption{Visual comparison results on the collected unpaired real-world rainy images using Google search with ``real rainy image".} 
	\label{fig:real_google}
\end{figure*}

We present five real-world test sets and some unpaired real-world rainy images using Google search with ``real rainy image" for qualitative comparison in Figs. \ref{fig:real_rd}-\ref{fig:real_google}. It is observed that our method outperforms existing methods when handling real-world data. This demonstrates that RainDiff can tackle various rain degradations in real-world rainy scenes and preserve image details effectively.
\begin{table}[h]
\scriptsize
 \renewcommand\arraystretch{1.0}
\centering
\caption{Quantitative results for ID, LIE and SR tasks.}
\label{com_app}
\setlength{\tabcolsep}{1.5mm}{
\begin{tabular}{c|c|c|c|c|c}
\toprule
\multirow{2}{*}{Model (ID)} & SIDD& \multirow{2}{*}{Model (LIE)} & LOL         & \multirow{2}{*}{Model (SR)} & DRealSR ($\times$ 4)  \\
 &PSNR/SSIM&                            & PSNR/SSIM   &                             & PSNR/SSIM   \\ \midrule
MM-BSN \cite{zhang2023mm}&37.37/0.936&EnlightenGAN \cite{jiang2021enlightengan}                & 17.48/0.651 & MetaKernelGAN \cite{lee2024meta}             & 31.06/0.834 \\
SCPGabNet \cite{lin2023unsupervised}&36.53/0.925&PairLIE \cite{fu2023learning}                   & 19.51/0.736 & ICF-SRSR \cite{neshatavar2024icf}                    & 30.65/0.821 \\
Ours                        & \textbf{39.28/}/\textbf{0.953} &Ours                        & \textbf{20.84}/\textbf{0.789} & Ours                 & \textbf{31.80}/\textbf{0.852} \\ \bottomrule
\end{tabular}}
\end{table}

\section{Extending to Other Image Translation Tasks.} 
The application scope of RainDiff is evidently broader than that of existing image deraining methods, as it possesses the capability to handle diverse unknown rain degradation types rather than being limited to a specific type. Moreover, the comparison with different image Denoising (ID), Low-light Enhancement (LIE), and Super-Resolution (SR) methods on SIDD, LOL, and DRealSR datasets is presented in Table \ref{com_app}, which demonstrates that our RainDiff possesses a broad application scope to handle different noise types and achieve superior performance in diverse image translation tasks.

\begin{table}[!h]
 \renewcommand\arraystretch{1.0}
\centering
\footnotesize
\caption{Quantitative results on the real-world test sets for different conditioning methods.}
\label{com_edit}

\begin{tabular}{c|c|c|c|c}
\toprule
\multirow{2}{*}{Model} & RS                                                                                  & RD                                                                                  & RDS                                                                                 & RHS                                                                                 \\
                       & PSNR/SSIM                                                                           & PSNR/SSIM                                                                           & PSNR/SSIM                                                                           & PSNR/SSIM                                                                           \\ \midrule
Refusion \cite{luo2023refusion}& 22.75/0.696                                                                         & 19.92/0.604                                                                         & 20.16/0.610                                                                         & 19.73/0.614                                                                         \\
Imagic \cite{kawar2023imagic}                  & 21.60/0.678                                                                         & 18.87/0.563                                                                         & 20.35/0.612                                                                         & 19.32/0.606                                                                         \\
InstructPix2Pix \cite{brooks2023instructpix2pix}      & 20.94/0.652                                                                        & 18.29/0.538                                                                         & 19.48/0.597                                                                        & 19.19/0.584                                                                         \\                    
Ours                   & \textbf{24.73}/\textbf{0.770} & \textbf{21.89}/\textbf{0.685} & \textbf{22.18}/\textbf{0.681} & \textbf{22.66}/\textbf{0.694} \\ \bottomrule
\end{tabular}
\end{table}

\section{Comparison with More Conditioning Methods} 
%
%
RainDiff can be regarded as an adaptation of existing conditioning diffusion models for real-world deraining with the ability to handle diverse real-world rain degradations in an unpaired learning manner, which cannot be achieved by previous conditioning ones. 
To show the superiority of RainDiff over conditioning methods, we compared RainDiff with five conditioning methods, including WeatherDiffusion \cite{ozdenizci2022restoring}, SR3 \cite{saharia2022image}, Refusion \cite{luo2023refusion}, and two image editing networks (Imagic \cite{kawar2023imagic} and InstructPix2Pix \cite{brooks2023instructpix2pix}) as exhibited in Table 2 (main draft), Table \label{ab_ddpm}, and Table \ref{com_edit}.

\begin{figure}[!t]
	\centering
	\includegraphics[width=0.8\linewidth]{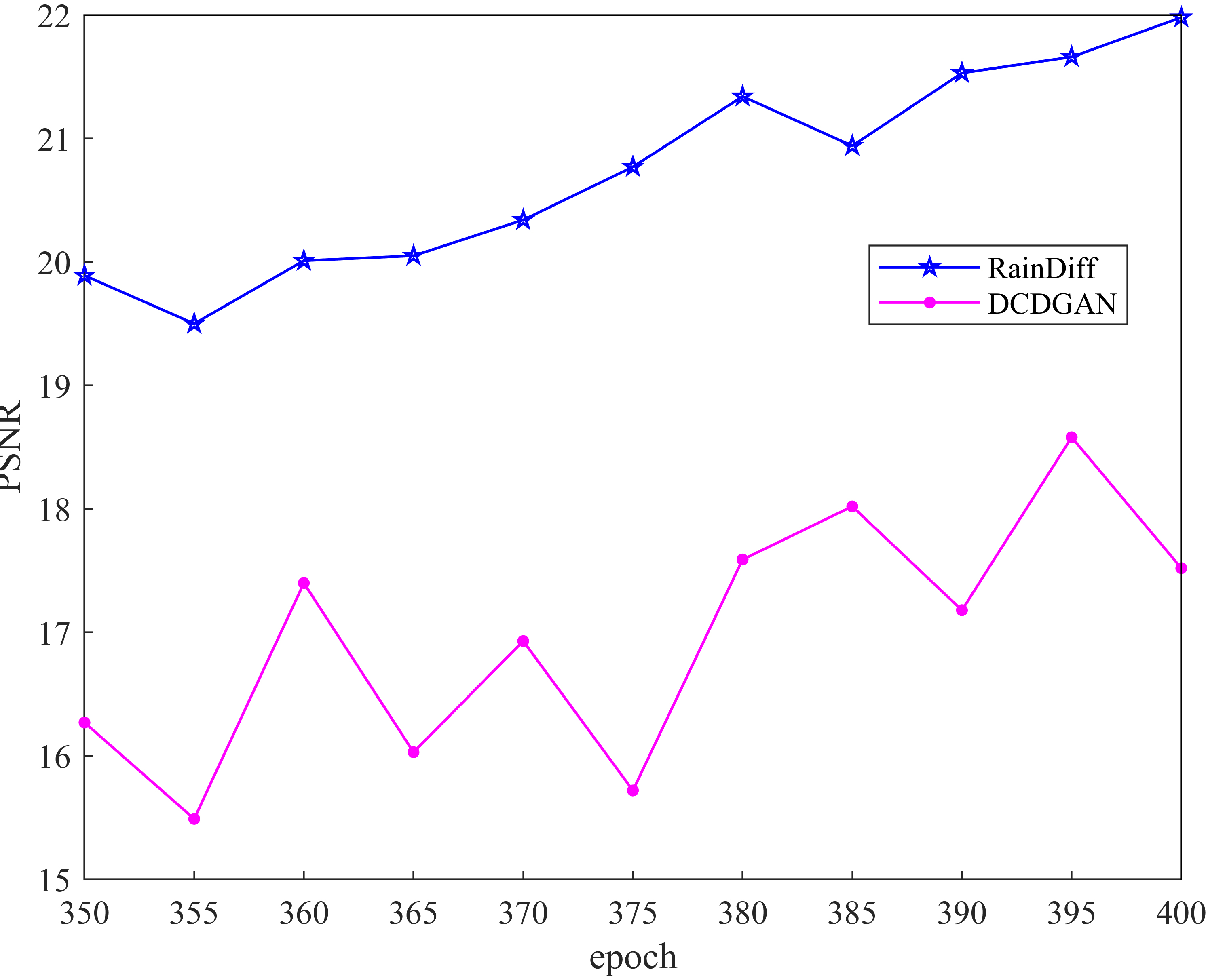}
	\caption{Analysis for stability of our method compared with GAN-based methods.} 
	\label{fig:stable}
\end{figure}

\section{Analysis for Stability}
We adopt a training process of 350-400 epochs to verify RainDiff's stability on the real-world RDS set compared with DCD-GAN. Moreover, RainDiff achieves superior performance with only 420 epochs, noticeably fewer than the 600 epochs required by DCD-GAN. It means that RainDiff provide more stable training process than GAN-based methods.

{
\bibliographystyle{splncs04}
\bibliography{main}
}